\documentclass[sigconf]{acmart}
\renewcommand\footnotetextcopyrightpermission[1]{} %
\newcommand{\method}{S\textsuperscript{2}-MoE}

\definecolor{ES}{RGB}{198, 219, 239}       
\definecolor{LS}{RGB}{158, 202, 225}       
\definecolor{EAGLE3}{RGB}{107, 174, 214}   
\definecolor{Cascade}{RGB}{49, 130, 189}   
\definecolor{S2-MoE}{RGB}{8, 81, 156}      
\usepackage{multirow}
\usepackage{bm}
\usepackage{bbding}
\usepackage{subfig}
\usepackage{adjustbox}
\usepackage{pgfplots}
\pgfplotsset{width=10cm,compat=1.16}

\begin{document}

\title{\method: Enabling Efficient Self-Speculative Decoding for Mixture-of-Experts on Edge Devices}

\newcommand{\affilHuang}{\textsuperscript{1,2}}
\newcommand{\affilQiu}{\textsuperscript{1,2,3}}
\newcommand{\affilLi}{\textsuperscript{1,2}}
\author{Haochen Huang\affilHuang}
\email{hhc@stu.pku.edu.cn}
\orcid{0009-0009-7863-5923}
\affiliation{%
  \institution{~}
  \city{~}
  \country{~}}

\author{Shengxuan Qiu\affilQiu}
\email{2300012877@stu.pku.edu.cn}
\affiliation{%
  \institution{~}
  \city{~}
  \country{~}}

\author{Meng Li\affilLi}
\authornote{Corresponding author.}
\email{meng.li@pku.edu.cn}
\orcid{0000-0002-7212-2264}
\affiliation{%
  \institution{~}
  \city{~}
  \country{~}}

\renewcommand{\shortauthors}{Huang et al.}

\begin{abstract}
Deploying large language models (LLMs) for inference on edge devices is challenging due to severe memory and bandwidth constraints. While speculative decoding and Mixture-of-Experts (MoE) have been proposed to improve inference efficiency, naively combining them often incurs excessive verification overhead and poor expert reuse, limiting their effectiveness in memory-bound edge settings. In this work, we propose \method, an efficient self-speculative decoding framework for MoE inference on edge devices. \method~reduces redundant verification through routing-aware adaptive speculative expansion, improves verification efficiency with reuse-aware expert gating, and aligns draft and target execution via shared context. Implemented in \texttt{llama.cpp}, \method~achieves up to \textbf{5.3$\times$} speedup (about \textbf{2.0$\times$} on average) over standard autoregressive decoding across diverse MoE models and datasets on edge devices. Code is available at \url{https://github.com/angerybob/S2-MoE}.
\end{abstract}




\maketitle
\pagestyle{fancy}
\fancyhead{} 
\begingroup
\renewcommand{\thefootnote}{}%
\footnotetext{%
  \textsuperscript{1}Institute for Artificial Intelligence, Peking University, Beijing, China\\
  \textsuperscript{2}School of Integrated Circuits, Peking University, Beijing, China\\
  \textsuperscript{3}School of Electronics Engineering and Computer Science, Peking University, Beijing, China}%
\endgroup

\section{Introduction}
\label{sec:intro}

Large language models (LLMs) have achieved remarkable performance across a wide range of tasks. However, many privacy-sensitive and latency-critical applications require LLM inference to be performed directly on edge devices. In such edge settings, the rapidly growing scale of LLMs poses significant challenges due to the limited resources available on edge platforms~\citep{edge,huang2025hd,fu2025h,xu2024edgellm,liu2023online}. 

A key bottleneck in edge LLM inference is memory bandwidth~\citep{memory}.
Edge inference workloads typically operate with small batch sizes and often degenerate to single-batch execution, resulting in limited parameter reuse and severe memory bandwidth pressure.

Under this setting, throughput is largely determined by how many tokens can be generated per access to model parameters.
This efficiency is influenced by two intuitive factors: the number of tokens generated per iteration, referred to as \textbf{output density}, and the amount of model parameters accessed per iteration, referred to as \textbf{parameter density}, leading to the following decomposition:
\begin{align}
& \text{Throughput}=\frac{\text{Output Density (tokens/iter.)}}{\text{Parameter Density (GB/iter.)}}\cdot \text{BW}\label{eq:throughput}
\end{align}
where the bandwidth $\text{BW}$ is limited in resource-constrained devices.

Building on this decomposition, existing edge inference acceleration techniques can be broadly categorized according to how they optimize these two factors.
On one hand, \textbf{speculative decoding} (SD)~\citep{he2024rest,speculative,miao2024specinfer,li2025eagle,zhong2024propd} accelerates autoregressive inference by first generating draft tokens with a lightweight draft model and then verifying them in parallel with the full model (often referred to as the target model). By allowing multiple tokens to be produced within a single iteration, speculative decoding effectively increases the output density.
On the other hand, \textbf{model sparsification}~\citep{zhou2022moe}, such as \textbf{Mixture-of-Experts} (MoE), reduces inference cost by activating only a subset of model parameters per iteration. This selective execution directly reduces the parameter density.

\begin{figure}[!t]
    \vspace{-6pt}
    \centering
    \includegraphics[width=\columnwidth]{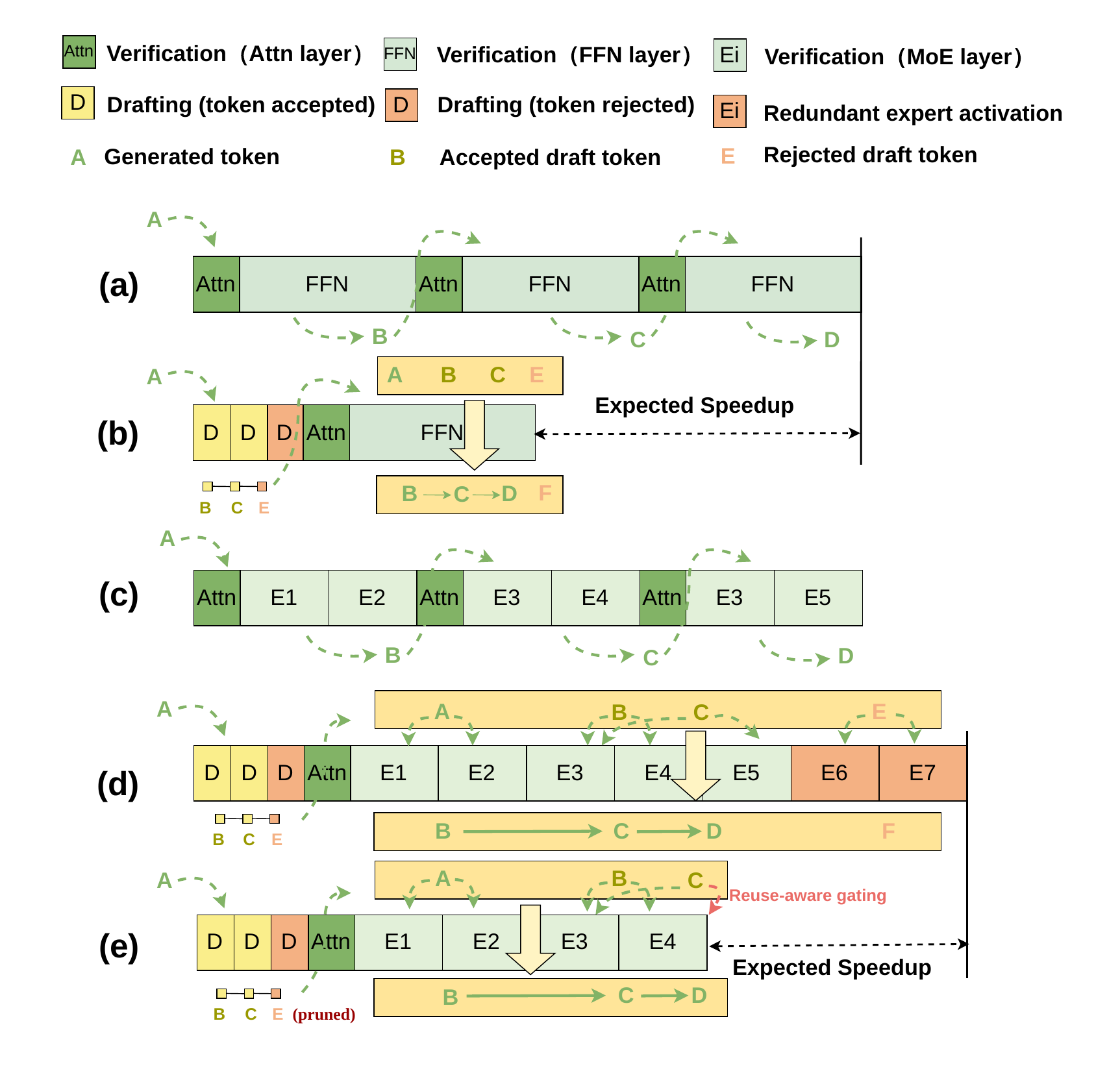}
    \vspace{-0.8cm}
    \caption{Comparison of autoregressive and speculative decoding on dense and MoE models. (a) Autoregressive decoding on a dense model. (b) SD on a dense model, where draft tokens can be efficiently verified due to parameter reuse across tokens. (c) Autoregressive decoding on a MoE model. (d) Naïve SD on a MoE model, where draft tokens activate different experts, leading to low parameter reuse and high verification cost. (e) \method~ improves SD on MoE models by reducing redundant verification and improving parameter reuse.}
    \label{fig:intro}
    \vspace{-10pt}
\end{figure}

Despite their complementary benefits, SD and MoE are difficult to combine effectively in edge inference settings. As illustrated in Figure ~\ref{fig:intro}, speculative decoding relies on high parameter reuse during verification.
While this property naturally holds for dense models, it breaks down for MoE models, where draft tokens may activate diverse experts, resulting in significantly increased verification cost.
This weakens the advantage of speculative decoding in memory-bound regimes. More critically, MoE models are highly sensitive to verification cost, since draft tokens that are ultimately rejected may still trigger disjoint expert activations, making verification disproportionately expensive. As a result, naïvely applying speculative decoding to MoE models can negate, or even reverse, the efficiency gains of either technique when used in isolation.

Most state-of-the-art SD methods improve throughput by generating and jointly verifying a large set of draft tokens to achieve sufficient acceptance length.
Representative approaches such as EAGLE-3~\citep{li2025eagle} rely on confidence-guided expansion and pruning to control speculative width.
While such strategies can suppress some low-quality branches, they remain difficult to calibrate in MoE settings.
Token-level confidence does not reflect expert-level verification cost, and even draft candidates with similar confidence may incur substantially different verification overhead since they may trigger disjoint expert activations.
Moreover, existing draft models are typically designed to be lightweight to keep drafting overhead low, but this often limits predictive fidelity, making it difficult to simultaneously achieve long acceptance lengths and low verification cost.

In this context, \textbf{self-speculative decoding}~\citep{self2,D&V,self3,anifmoe,zeng2025self,quant} offers a unique opportunity to better align draft and target behaviors for speculative decoding in MoE models. Instead of relying on an external draft source, prior self-speculative approaches construct a lightweight draft directly from the target model itself. Because the draft is derived from the same underlying model, its prediction behavior is naturally better aligned with that of the target. This stronger alignment makes self-speculative decoding a more promising starting point for achieving long acceptance lengths with a lightweight draft, while also reducing redundant verification.

While self-speculative decoding offers strong potential for aligning draft and target behaviors, it is not directly applicable to MoE models, as achieving sufficient lightweight drafts inevitably introduces fidelity loss. First, imperfect alignment between draft and target still provides limited acceptance rates. Second, rejected drafts lead to excessive redundant verification, eroding the benefits of speculative decoding. Third, MoE models inherently exhibit limited expert parameter reuse, and diverse expert activations across speculative tokens further exacerbate this issue, significantly increasing verification cost. Therefore, a design that simultaneously ensures high draft fidelity, low redundant verification, and strong parameter reuse is required for SD in MoE models.

To meet this requirement, we propose \textbf{\method}, an efficient self-speculative decoding framework for MoE inference on edge devices (Figure~\ref{fig:intro}(e)). In summary, we contribute:
\textbf{(1) Routing-aware Adaptive Speculative Expansion}, which dynamically expands speculative candidates based on their \emph{verification utility}, allowing the system to avoid low-quality candidates with high verification cost and thereby reducing redundant verification;
\textbf{(2) Reuse-aware expert gating}, a MoE gating design that explicitly promotes expert parameter reuse across draft tokens, thereby increasing the computational density of verification;
\textbf{(3) Context-aligned self-speculative decoding}, which enables the lightweight draft and the target model to \emph{share the same decoding context KV Cache}, preventing approximation errors in the draft from accumulating across iterations and thereby improving acceptance rates;
(4) A full system implementation integrated into \textbf{llama.cpp}, demonstrating end-to-end efficiency gains. Across diverse MoE models and datasets, \method\ achieves 1.3$\times$ to 5.3$\times$ speedup on Jetson Orin and 1.2$\times$ to 2.9$\times$ on RTX~4090 over standard autoregressive decoding. Our code is available at \url{https://github.com/angerybob/S2-MoE}.

\section{Background}
\label{sec:background}



\subsection{Mixture-of-Experts}

Mixture-of-Experts (MoE) models improve scalability by sparsely activating a small subset of experts per token, reducing effective computation and parameter access compared to dense models.
This sparse execution enables large-capacity models to operate under lower average memory and compute cost, making MoE architectures attractive for memory-constrained inference scenarios, including edge LLM deployment~\citep{zhou2022moe,deepseekv3, deepseekv2,openai2025gptoss120bgptoss20bmodel,qwen,yang2024qwen2technicalreport,qwen3technicalreport}.

\subsection{Speculative Decoding}


Speculative decoding (SD)~\citep{he2024rest,speculative,miao2024specinfer,li2025eagle,zhong2024propd} accelerates autoregressive decoding by using a lightweight mechanism to propose future tokens and letting the target model verify them in parallel, which is particularly beneficial in memory-bound settings. Existing SD methods mainly differ in how speculative candidates are generated, including methods based on an external draft model, methods using internal features or auxiliary prediction heads, and parallel decoding methods without an explicit draft.

For MoE models, however, these approaches remain difficult to apply effectively. 
External-draft methods require a draft model that is both lightweight and well aligned with the target~\citep{speculative,miao2024specinfer}, which is difficult to achieve in practice. This challenge is even more pronounced for MoE models, where only a small subset of parameters is already activated per token, leaving limited room to construct a substantially smaller yet well-aligned draft.
Methods based on auxiliary heads or parallel decoding~\citep{he2024rest,li2025eagle,zhong2024propd} avoid an external draft, but typically still require generating and verifying many speculative candidates to maintain a useful acceptance length. In MoE models, this can be especially expensive because different candidates may activate different experts, leading to high and redundant verification overhead. 
Although confidence-based expansion and pruning (e.g., EAGLE-3~\citep{li2025eagle}) can reduce some low-quality branches, token-level confidence does not directly reflect expert-level verification cost. Moreover, the effectiveness of such predictors often depends heavily on the model family and predictor training quality, which can make it difficult to consistently sustain long acceptance lengths and large speedups across MoE models.

\subsection{Self-Speculative Decoding}


Motivated by this gap, self-speculative decoding eliminates the need for an external draft by deriving a lightweight draft directly from the target model itself,
thereby achieving strong behavioral alignment.
Existing self-speculative approaches typically construct such drafts through various forms of model simplification. Common strategies include quantization~\citep{quant}, expert sparsity~\citep{anifmoe}, which activates fewer experts per token in MoE models, and layer sparsity~\citep{D&V,self2,self3}, which skips selected layers or enables early exit during decoding. These techniques form a spectrum of trade-offs: while more aggressive simplification reduces draft execution cost, it also degrades prediction quality and limits acceptance rates. 

\subsection{Speculative Decoding for MoE Models}
Existing studies have explored applying SD to MoE models from multiple perspectives. 
Some works leverage SD to predict future expert activations, enabling expert prefetching and offloading, but primarily focus on system-level caching and scheduling without addressing the intrinsic inefficiencies of verification under sparse activation~\citep{chen2025sp,zhuge2025specoffload,wang2025moe,wang2025accelerating,zheng2026ssmoe,li2026moespac,bang2026specmoe}. 
Other efforts observe that SD is not universally beneficial for MoE inference and propose utility- or cost-aware policies to dynamically enable speculation and adjust the speculation length. 
However, such approaches are driven by historical utility signals, whose reliability can degrade under unstable routing, limiting their ability to consistently improve speculative efficiency~\citep{saxena2025utility}.
More recently, SD has been shown to provide speedups for MoE models under moderate batch sizes by amortizing verification overhead across requests, yet these benefits largely diminish in edge inference settings, where batch sizes are typically one or two and verification remains memory-bound~\citep{huang2025moesd}. Importantly, these studies primarily provide empirical and analytical insights into this behavior, rather than addressing the underlying inefficiencies.

\begin{figure}[!t]
    \vspace{-6pt}
    \centering
    \includegraphics[width=\columnwidth]{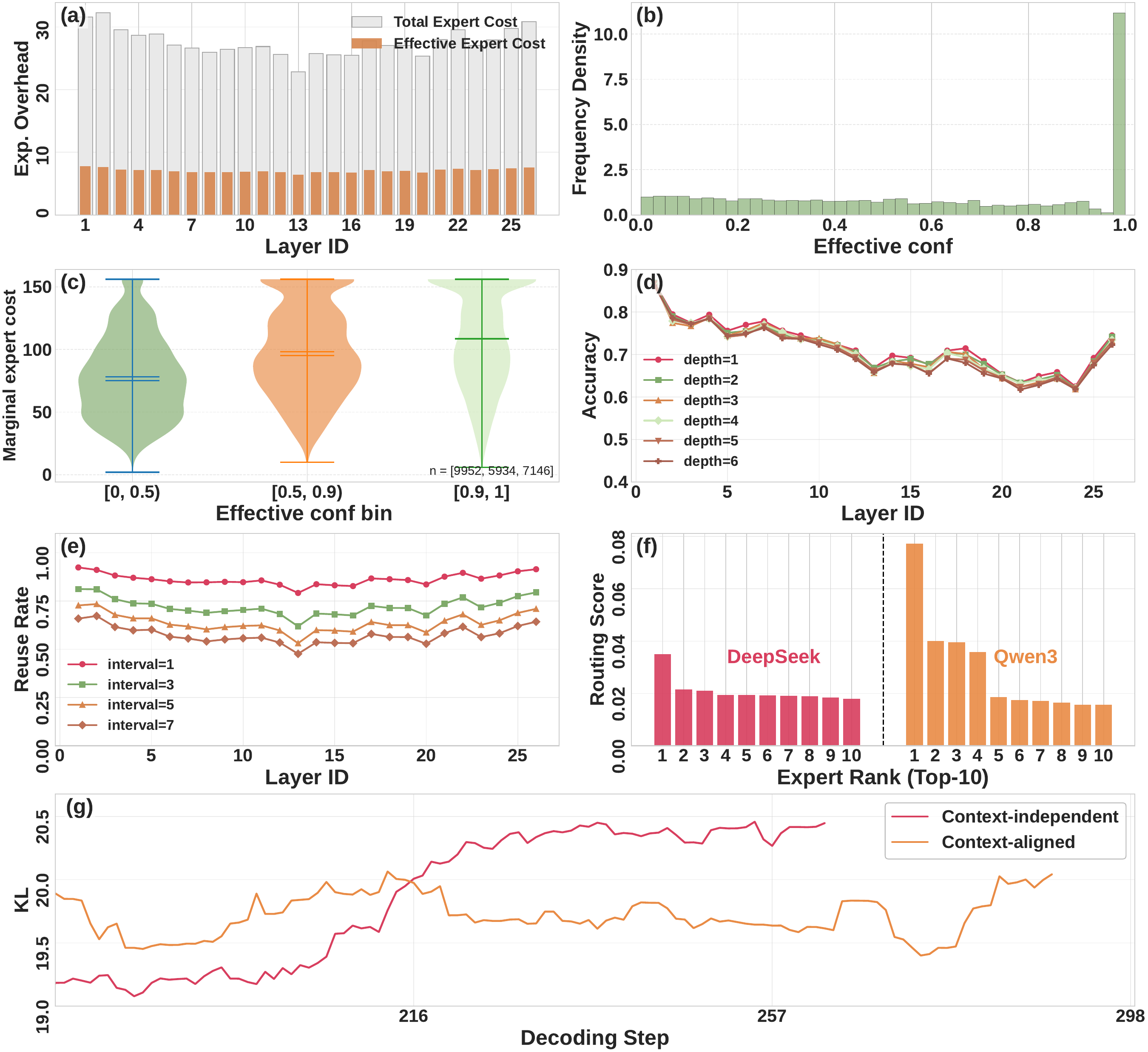}
    \vspace{-6pt}
    \caption{(a) Layer-wise breakdown of verification cost for DeepSeekMoE evaluated in Natural Questions, highlighting redundant expert activation. (b) Distribution of Draft Confidence. (c) Verification Cost Distribution Across Confidence Levels. (d) Accuracy of draft expert routing in predicting target routing under different draft depths.  (e) Expert reuse ratio across layers under different decoding intervals. (f) Distribution of top routing scores, revealing expert redundancy. (g) KL divergence between draft and target models over decoding steps w/ and w/o shared context. }
    \label{fig:motivation}
    \vspace{-4pt}
\end{figure}

\section{Motivation}
\label{sec:motivation}
\begin{figure}[!t]
    \vspace{-4pt}
    \centering
    \includegraphics[width=0.95\columnwidth]{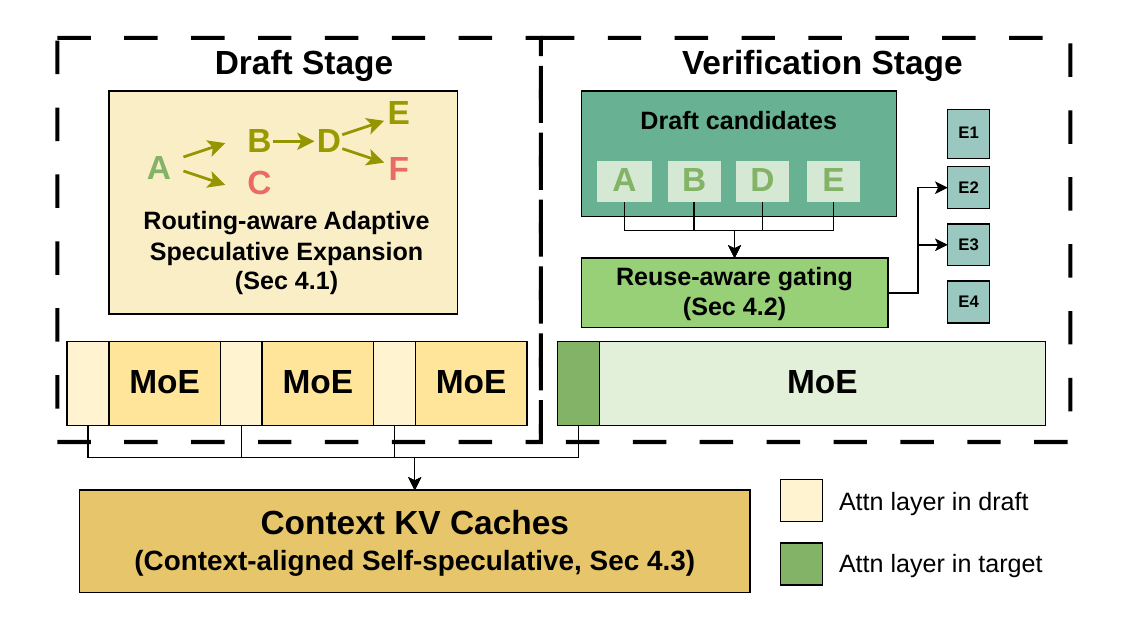}
    \vspace{-15pt}
    \caption{\method~overview.}
    \label{fig:overview}
    \vspace{-15pt}
\end{figure}

\label{sec:method}
\begin{figure*}[!t]
    \vspace{-4pt}
    \centering
    \includegraphics[width=2.1\columnwidth]{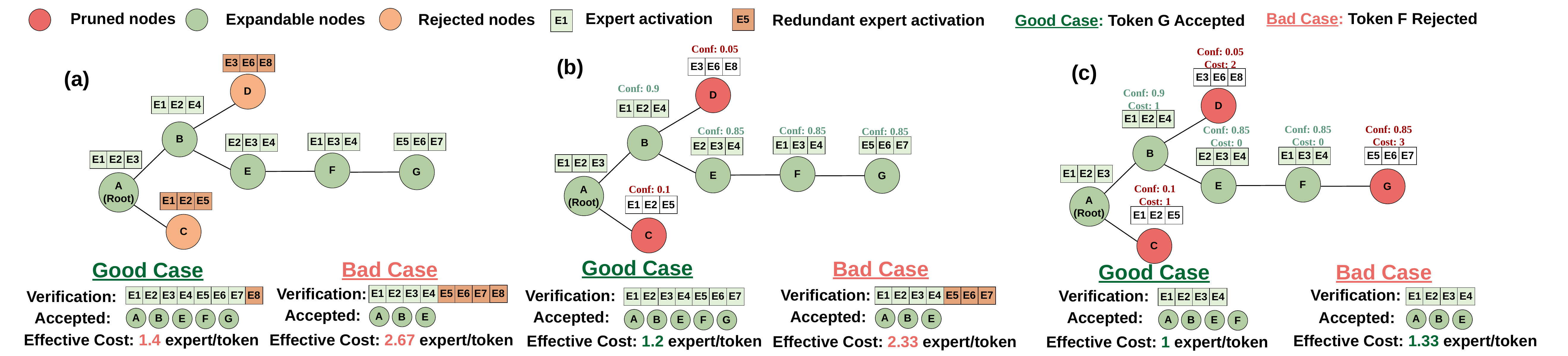}
    \vspace{-20pt}
    \caption{Routing-aware adaptive expansion reduces verification cost under different acceptance outcomes. (a) Expansion without pruning. (b) Confidence-based pruning. (c) Our routing-aware adaptive expansion.}
    \label{fig:ae}
    \vspace{-0.3cm}
\end{figure*}

\textbf{Challenge 1: Excessive redundant verification under low-fidelity drafts.} 
In MoE models, verifying each draft token may activate additional experts, incurring nontrivial parameter access.
When draft tokens are rejected, these extra activations become redundant verification cost.
Fig.~\ref{fig:motivation}(a) shows that, across layers, a large fraction of verification cost is spent on experts triggered by ultimately rejected tokens, highlighting the inefficiency of naïve speculation.
Moreover, this redundancy is difficult to eliminate with confidence-only pruning. As shown in Fig.~\ref{fig:motivation}(b), draft confidence is highly concentrated near 1.0 for retained tokens, making confidence weakly discriminative in practice. At the same time, Fig.~\ref{fig:motivation}(c) shows that these tokens can still incur substantially different verification costs in MoE models due to different expert activation patterns. This motivates speculative expansion that jointly considers acceptance potential and verification overhead.

\textbf{Opportunity 1: Predicting verification cost via routing similarity.} 
As illustrated in Fig.~\ref{fig:motivation}(d), draft-time expert routing predicts target routing with high accuracy in many layers, enabling draft routing signals to serve as a proxy for estimating verification cost.
This motivates adaptive speculative expansion that explicitly balances potential acceptance benefit (typically estimated from the draft’s token-level confidence as in prior work) against expected verification overhead.

\textbf{Challenge 2: Limited expert parameter reuse under diverse expert activations.} 
Unlike dense models, where adjacent tokens share the same parameters, MoE models often activate different experts for neighboring tokens.
We measure expert reuse using a \textit{reuse ratio}, defined for a token interval of length $s$ as the number of unique experts activated within that interval during parallel verification, normalized by the number of expert activations in autoregressive decoding over the same interval. A smaller reuse ratio indicates stronger cross-token expert reuse. In the ideal case where all tokens within the interval reuse the same set of experts, the ratio approaches $1/s$, which corresponds to a dense-like parameter reuse pattern across tokens.
As shown in Fig.~\ref{fig:motivation}(e), this ratio remains high across layers under different decoding intervals, indicating limited expert reuse and substantially higher parameter traffic under parallel verification.

\textbf{Opportunity 2: Flexible routing among low-impact experts.} 
In MoE models, expert importance is often skewed, with a few high-impact experts receiving dominant gating scores, while others receive lower scores. Experts with similar low scores, though not selected, can often perform similarly. 
Fig.~\ref{fig:motivation}(f) shows that beyond the top-ranked experts, many candidates exhibit comparable routing scores, indicating functional redundancy. 
This creates an opportunity to trade a small loss in per-token gating score for increased expert reuse, by allowing a token to select experts with slightly lower scores that are already activated by other draft tokens.

\textbf{Challenge 3: Limited draft fidelity under lightweight draft.} 
Although self-speculative drafts are derived from the target model itself, lightweighting still introduces prediction errors. When the draft maintains an independent context, these errors accumulate over decoding steps, causing progressive divergence and reducing acceptance rates in long-form generation. As shown in Fig.~\ref{fig:motivation}(g), the KL divergence between draft and target distributions grows rapidly in this case when the draft maintains an independent context, indicating severe error accumulation over time.

\textbf{Opportunity 3: Reducing error accumulation via shared context.} 
In autoregressive language models, next-token prediction follows $P(y_t \mid x, y_{<t})$, so even small errors in the conditioning context can affect subsequent predictions and accumulate over decoding steps. By sharing the same decoding context between the draft and target, such errors are confined to the current speculative step instead of propagating across iterations. As shown in Fig.~\ref{fig:motivation}(g), context alignment substantially suppresses divergence growth over long contexts.

\section{\method~ Design}

Figure~\ref{fig:overview} illustrates the overall design of \method. Our framework combines three complementary components to improve speculative decoding for MoE models on edge devices: routing-aware adaptive speculative expansion (Sec.~\ref{sec:method_utility}), reuse-aware expert gating (Sec.~\ref{sec:method_reuse}), and context-aligned self-speculative decoding (Sec.~\ref{sec:method_context_aligned}). We next describe these components in turn.

\begin{figure*}[!t]
    \vspace{-4pt}
    \centering
    \includegraphics[width=2\columnwidth]{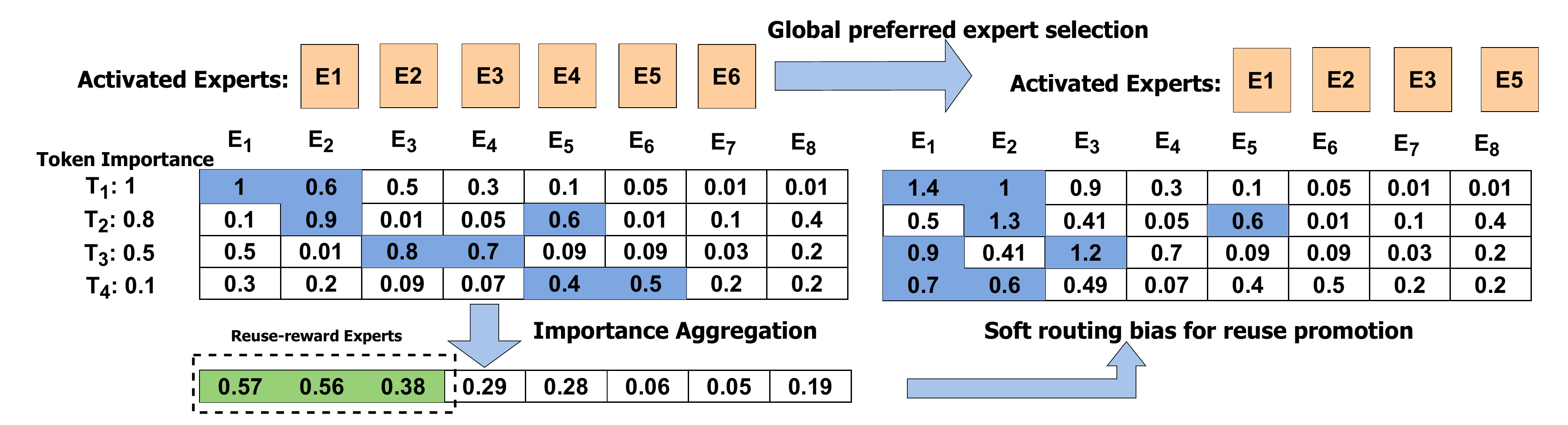}
    \vspace{-6pt}
    \caption{An illustrative example of reuse-aware expert gating with top-2 routing.
Four draft tokens select two experts each from eight experts.
Token-level confidence is used to aggregate expert importance and identify a small
set of reuse-reward experts ($E_1$–$E_3$), whose logits are softly biased to improve
cross-token expert reuse during verification.}
    \label{fig:reuse}
    \vspace{-0.5cm}
\end{figure*}

\subsection{Routing-aware Adaptive Speculative Expansion}
\label{sec:method_utility}

To reduce redundant verification in MoE speculative decoding, we adaptively expand the draft tree according to a utility score that jointly accounts for acceptance likelihood and verification cost.
This design is motivated by a key limitation of prior adaptive expansion rules:
in dense models, token confidence is often a reasonable proxy for speculative utility,
but in MoE models, the verification cost depends heavily on which experts are activated.
As a result, two draft candidates with similar confidence can induce substantially different verification overhead if one largely reuses already activated experts while the other introduces many new ones.

\paragraph{Expected Benefit.}
Following prior speculative decoding works, for a draft token $i$ at depth $k$, we estimate its potential acceptance probability as the prefix confidence $p_i = \prod_{j=1}^{k} p_{i,j}$,
since token $i$ can be accepted only if all preceding draft tokens are also accepted.
If accepted, token $i$ saves one autoregressive decoding step of the target model.
We therefore define its expected benefit as $B_i = p_i \cdot T_{\mathrm{AR}}$, where $T_{\mathrm{AR}}$ denotes the average latency of one target-model decoding step.

\begin{figure}[t]
    \centering
    \includegraphics[width=\columnwidth]{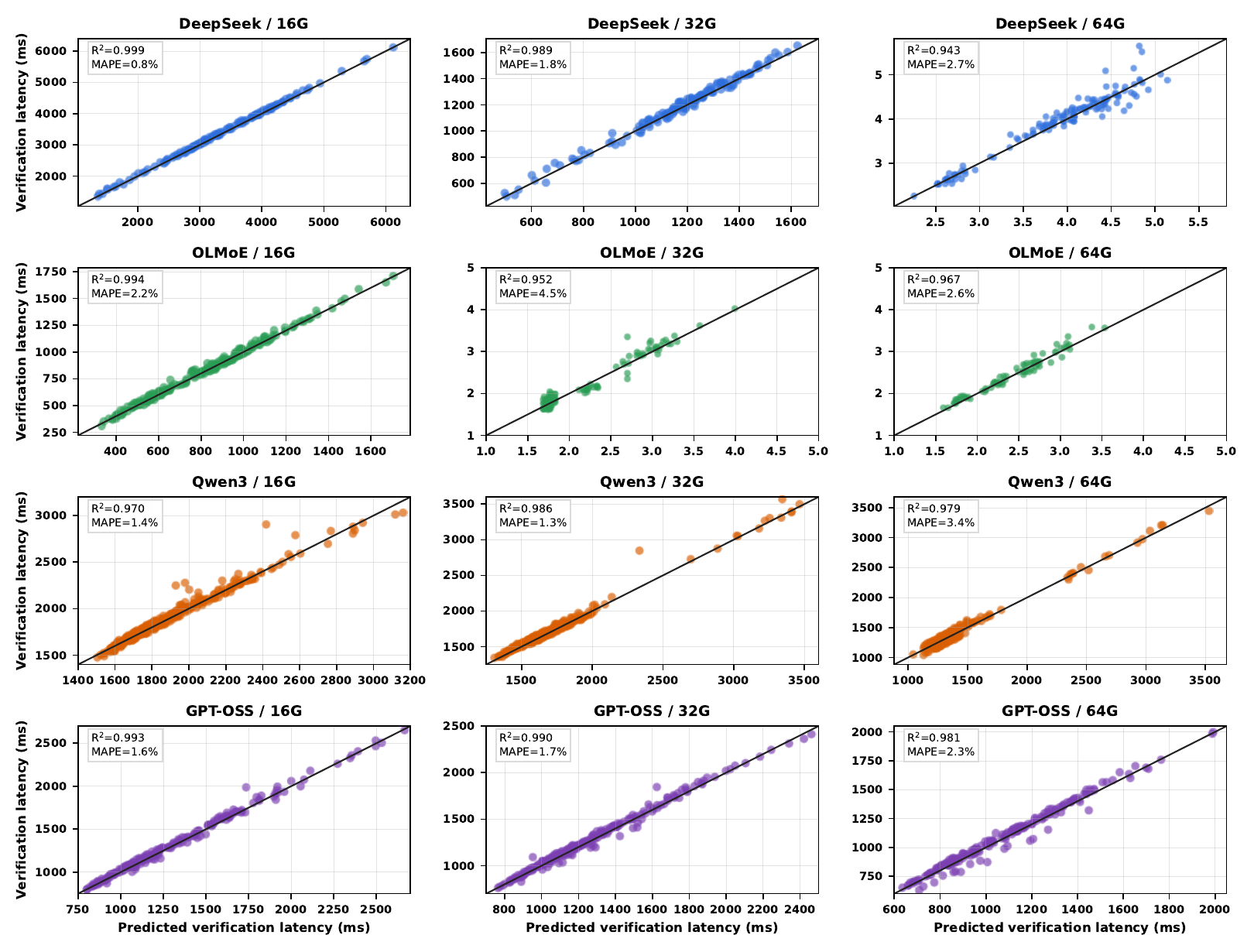}
    \caption{Cost-model calibration against measured verification latency across models and memory budgets.}
    \label{fig:cost_model}
\end{figure}

\paragraph{Routing-aware Cost Estimation.}
The cost of expanding token $i$ consists of two parts:
(i) the draft expansion cost for generating its descendants, and
(ii) the verification cost incurred when validating it with the target MoE model.
Unlike dense models, where verification cost increases little with the number of draft tokens, MoE verification cost increases significantly as more tokens are validated, since each additional token may activate a new set of experts.
Let $\mathcal{E}_i$ denote the predicted experts for token $i$, and let $S$ denote the set of draft tokens already selected for expansion. We define the marginal verification cost term of adding token $i$ as
\begin{equation}
\widehat{\Delta C}^{\mathrm{ver}}(i \mid S)
=
\left|
\mathcal{E}_i \setminus \bigcup_{j \in S} \mathcal{E}_j
\right| \cdot T_{\mathrm{exp}},
\end{equation}
where $T_{\mathrm{exp}}$ denotes the latency associated with introducing one additional expert during verification and can be obtained through lightweight runtime measurement or offline profiling. This term converts newly introduced experts into a hardware-calibrated latency unit, enabling direct comparison between verification cost and expected benefit.
This estimate uses draft-time routing signals, which are available before verification and correlate with target routing across layers, as demonstrated in Opportunity 1.
As shown in Fig.~\ref{fig:cost_model}, the resulting cost model closely tracks measured verification latency across models and memory budgets ($R^{2}{=}0.943$--$0.999$, MAPE${=}0.8$--$4.5\%$), indicating that it reliably predicts the actual verification overhead used by adaptive expansion.
The total expansion cost is then defined as
$C_i = \widehat{\Delta C}^{\mathrm{ver}}(i \mid S) + C_i^{\mathrm{draft}}$.

\paragraph{Adaptive Expansion.}
The utility of a draft token $i$ is defined as

\begin{equation}
U_i = \frac{B_i}{C_i}
    = \frac{p_i \cdot T_{\mathrm{AR}}}
           {\widehat{\Delta C}^{\mathrm{ver}}(i \mid S) + C_i^{\mathrm{draft}}}.
\end{equation}

Speculative expansion proceeds greedily by admitting candidates with $U_i \ge 1$, i.e., when the expected benefit exceeds the estimated cost, until no further candidates satisfy the criterion.

Compared with fixed-width or confidence-only expansion, our routing-aware utility captures a MoE-specific effect that prior methods overlook: under MoE verification, candidate quality depends not only on acceptance likelihood but also on marginal expert activation cost.
As illustrated in Fig.~\ref{fig:ae}, expansion without pruning (a) verifies all draft candidates, leading to substantial redundant expert activation, while confidence-based pruning (b) removes clearly low-quality branches but remains blind to expert-level verification cost and may still retain high-cost candidates in some cases.
In contrast, our routing-aware adaptive expansion (c) jointly considers acceptance likelihood and marginal verification cost, favoring candidates with better acceptance-to-verification trade-offs and reducing redundant expert activation.
This advantage holds even when acceptance outcomes are unfavorable (Bad Case in Fig.~\ref{fig:ae}), as prioritizing low-cost candidates still limits redundant overhead.

\subsection{Reuse-aware Expert Gating}
\label{sec:method_reuse}


In contrast to dense models, where neighboring tokens reuse parameters, MoE models often route consecutive tokens to different experts.
This effect is amplified under speculative decoding with parallel verification, resulting in heavy expert activation and reduced parameter reuse, which increases memory overhead.

To address this problem, we introduce \textbf{reuse-aware expert gating}, a gating mechanism that softly aligns expert routing decisions across speculative tokens.
The core idea is to first identify experts that already exhibit repeated activation across a verification batch,
and then softly amplify this existing reuse tendency,
thereby improving expert parameter locality and reducing memory traffic.

\paragraph{Cross-token expert importance aggregation.}

Tokens differ in their likelihood of being accepted during verification.
High-confidence tokens are more likely to be accepted and therefore should preserve their original expert preferences,
whereas low-confidence tokens are less likely to survive during verification and mainly contribute to redundant computation.
To reflect this asymmetry, we associate each token with a non-negative confidence weight $w_b$,
derived from its token-level confidence, so that expert preferences of high-confidence tokens receive greater emphasis in aggregation.

Using these confidence weights, we consider a set of draft tokens
$\mathcal{B} = \{1,\dots,B\}$ and an MoE layer with expert set
$\mathcal{E} = \{1,\dots,E\}$.
For each token $b \in \mathcal{B}$, let $\ell_b \in \mathbb{R}^E$ denote
the expert gating logits produced by the target model.
We aggregate expert importance across speculative tokens as

\begin{equation}
g_e = \sum_{b \in \mathcal{B}} w_b \cdot \ell_{b,e}, \qquad \forall e \in \mathcal{E},
\end{equation}

where $g_e$ captures how strongly expert $e$ is favored by speculative tokens
that are more likely to be accepted.
This aggregation jointly accounts for token-level confidence and expert preference,
thereby identifying experts that are most valuable to prioritize for reuse during verification.

\paragraph{Global preferred expert selection.}
To avoid over-concentration and preserve the original top-$k$ routing structure,
We restrict biasing to a bounded set of experts.
Specifically, we select a set of \emph{globally preferred experts} $\mathcal{E}^\star = \operatorname{TopCap}(\{g_e\}_{e=1}^E)$,
where $|\mathcal{E}^\star| \le \texttt{cap}$ and $\texttt{cap} \ge k$.
This ensures that all originally eligible top-$k$ experts remain selectable,
while limiting the scope of routing bias.

\paragraph{Soft routing bias for reuse promotion.}
For every speculative token $b$, we inject an additive bias into its gating logits $\ell'_{b,e} = \ell_{b,e} + \lambda_b \cdot \mathbf{1}[e \in \mathcal{E}^\star]$, where $\lambda_b$ is determined based on the routing score distribution of token $b$.
Specifically, we set $\lambda_b$ proportional to the gap between the top-1 expert and the $(k+1)$-th expert.
This gap provides a natural measure of the margin for entering the top-$k$ set, allowing the bias to promote reuse-reward experts into the selection boundary without overly perturbing dominant routing decisions.
The required routing scores are readily available from the gating module during inference, and can also be obtained through lightweight offline profiling, introducing no additional training requirement.
In practice, the bias primarily affects borderline experts near the top-$k$ threshold, while preserving the relative ordering of strongly preferred experts, thereby improving expert reuse without significantly degrading routing quality. 
Moreover, because this bias increases expert overlap across draft tokens, the realized verification cost is generally no higher than that estimated from the original draft routing, making the cost estimate in Sec.~\ref{sec:method_utility} conservative.



\paragraph{Efficient fused implementation.}
We implement reuse-aware expert gating as a lightweight fused CUDA kernel, introducing negligible overhead during speculative decoding.

Figure~\ref{fig:reuse} illustrates reuse-aware expert gating with a toy example.
Routing information from four draft tokens with different confidence levels is aggregated to identify a small set of reuse-reward experts (e.g., $E_1$–$E_3$), whose gating logits are softly biased.
This bias reduces the number of activated experts during verification (from six to four in this example) while remaining token-adaptive:
high-confidence tokens (e.g., $T_1$ and $T_2$) largely retain their original expert choices, whereas low-confidence tokens (e.g., $T_4$) are more likely to shift toward reuse-reward experts.

\subsection{Context-aligned Self-speculative}
\label{sec:method_context_aligned}




A major source of fidelity loss in self-speculative decoding is context misalignment between the draft and target models, where independently maintained decoding states cause errors to accumulate over time. To address this issue, we propose context-aligned self-speculative decoding, in which the draft and target share a unified KV cache.

Specifically, the target model first processes the prompt and establishes the KV cache. During speculative generation, the draft directly reads from this shared context instead of maintaining a separate history, eliminating context drift. To preserve correctness, KV updates from the draft are treated as temporary and rolled back before verification, and only accepted tokens are committed to the shared cache.

This design ensures that draft errors are confined to individual speculative steps and do not accumulate across iterations.


\section{System Implementation}


\paragraph{Framework.}
We implement our approach on top of \textbf{llama.cpp}, a widely used inference framework
for edge and resource-constrained deployment.
All proposed mechanisms in \method~are fully integrated
into the existing decoding pipeline.

\paragraph{Edge memory constraints and expert-level offloading.}
Edge platforms often operate under tight memory budgets, especially when serving
large MoE models.
We implement expert-level offloading on both NVIDIA Jetson Orin and a discrete-GPU
NVIDIA RTX~4090 platform.
On Jetson Orin, the CPU and GPU share a unified DRAM without dedicated host memory,
so parameters exceeding device memory are offloaded to lower-tier storage such as SSDs.
On RTX~4090, with separate CPU/GPU memories, cold experts are offloaded to CPU memory
and transferred on demand.

While \textbf{llama.cpp} natively supports layer-level parameter offloading,
this granularity is ill-suited for MoE models because every layer is visited during inference, but only a small subset of experts
is activated per token.
Offloading entire layers therefore incurs unnecessary data movement.
To address this mismatch, we implement expert-level parameter offloading,
where non-expert and shared parameters remain resident in device memory,
and individual expert parameters are dynamically fetched on demand. When memory permits, we further keep frequently activated experts resident
in device memory based on offline profiling, while ensuring sufficient
runtime memory headroom for online inference.

Our work focuses on improving speculative decoding efficiency and is orthogonal
to prior efforts on parameter offloading and caching strategies.
Existing offloading methods primarily optimize data movement through prefetching
and cache management, and their ideal operating regime corresponds to scenarios
with high cache hit rates.
Such settings are effectively equivalent to memory-relaxed configurations,
which we explicitly evaluate (e.g., 64GB memory budgets).

Moreover, to the best of our knowledge, no existing MoE offloading work provides a publicly available implementation compatible with \textbf{llama.cpp}. Re-implementing these systems would require substantial engineering effort beyond the scope of this work.

\paragraph{Baseline models and implementations.}
We evaluate \method\ against state-of-the-art SD baselines
implemented on representative MoE model families.
For GPT-OSS, Qwen3, and OLMoE models, we use publicly released EAGLE-3 checkpoints
from Hugging Face, including lmsys/EAGLE3-gpt-oss-120b-bf16~\citep{hf_eagle3_gptoss},
AngelSlim/Qwen3-a3B\_eagle3~\citep{hf_qwen3_eagle3}, and wantsleep/OLMoE\_1B\_7B\_Eagle3~\citep{wantsleep_olmoe_eagle3}.
For DeepSeek models, EAGLE-3 style speculative decoding is implemented based on
the open-source SpecForge framework built on sglang~\citep{specforge}.
All EAGLE-3 baselines are executed using the official EAGLE-3 implementation
in \textbf{llama.cpp}, based on the corresponding development branch
(see PR~\#18039)~\citep{llamacpp_eagle3}, which we adapt and integrate into our evaluation pipeline. For Cascade~\citep{saxena2025utility}, we implement its core policy in \textbf{llama.cpp} following the method described in their paper, and integrate it into the same evaluation pipeline for a fair comparison.

\paragraph{Draft configuration.}
Our draft is constructed based on expert sparsity because quantization often incurs
substantial draft overhead, while aggressive layer sparsity can severely
degrade draft fidelity.
The detailed configurations are
summarized in Table~\ref{tab:model_draft_configs}.
When memory permits, we additionally apply quantization to the sparse draft to further reduce computation and memory footprint.
The draft length is fixed at 8, and the draft width is set to 2 for baseline models, while \method~ uses adaptive expansion based on utility scores.

\begin{table}[H]
\centering
\caption{Model and SD configurations.}

\label{tab:model_draft_configs}
\scriptsize
\setlength{\tabcolsep}{1pt}
\begin{tabular}{lccccccc}
\toprule
& \multicolumn{4}{c}{Model configuration} & \multicolumn{3}{c}{SD hyperparameter} \\
\cmidrule(lr){2-5} \cmidrule(lr){6-8}
Model & Total Params & Activated Params & Total Experts & Experts/Token & Cap & Draft Experts \\
\midrule
OLMoE    & 6.7   & 1.1  & 64  & 8 & 18& 2 \\
DeepSeek & 15.3  & 1.35 & 64  & 6 & 16& 1 \\
Qwen3    & 30.5  & 3.3  & 128 & 8 & 14& 3 \\
GPT-OSS  & 116.8 & 5.7  & 128 & 4 & 6 & 1 \\
\bottomrule
\vspace{-15pt}
\end{tabular}
\end{table}

\section{Experiments}
\label{sec:experiments}

\subsection{Experimental Setup}

\paragraph{Models.} We evaluate our method on four representative MoE-based LLMs covering diverse model scales and design generations: DeepSeek-V2-Lite-Chat (DeepSeek)~\citep{deepseekv2}, OLMoE-1B-7B (OLMoE)~\citep{muennighoff2025olmoeopenmixtureofexpertslanguage}, Qwen3-30B-A3B (Qwen3)~\citep{qwen3technicalreport}, and GPT-OSS-120B (GPT-OSS)~\citep{openai2025gptoss120bgptoss20bmodel}.
These models span from lightweight edge-friendly MoE models to large-scale, high-capacity MoE systems, enabling a comprehensive evaluation across different parameter sizes, expert configurations,
and routing behaviors (see Table~\ref{tab:model_draft_configs}).

\paragraph{Datasets and Tasks.}
Following EAGLE-3 and Spec-Bench~\citep{specbench}, we evaluate effectiveness across a diverse set of generation tasks, including multi-turn conversation (\textbf{MT}), retrieval-augmented generation (\textbf{RG}), summarization (\textbf{SU}), translation (\textbf{TR}), question answering (\textbf{QA}), mathematical reasoning (\textbf{MA}), and code generation (HumanEval, \textbf{HE}), using identical speculative decoding hyperparameters across all tasks for fairness.
For quality evaluation, we assess whether reuse-aware gating introduces degradation in model quality.
Specifically, DeepSeek and OLMoE are evaluated on GSM8K~\citep{cobbe2021trainingverifierssolvemath}, HumanEval~\citep{human-eval}, ARC-E, and ARC-C~\citep{arc}, while the stronger Qwen3 and GPT-OSS are evaluated on GPQA-Diamond~\citep{gpqa}, AIME24~\citep{AIME2024}, AIME25~\citep{AIME2025}, and HMMT~\citep{hmmt_2025}.
This split ensures that each model is evaluated on tasks aligned with its reasoning capability.
We additionally report long-context quality on GovReport, NarrativeQA, and Qasper from LongBench~\citep{bai2024longbench}, together with distribution-consistency metrics against the unmodified target on WikiText-2~\citep{merity2017pointer}.

\paragraph{Hardware Platform.}
We evaluate on two platforms: NVIDIA Jetson Orin edge devices and a discrete-GPU NVIDIA RTX~4090.
On Jetson Orin, we consider three memory constraints: 16~GB, 32~GB, and 64~GB, corresponding to Jetson Orin NX 16GB and Jetson AGX Orin 32GB/64GB.
When expert parameters exceed on-device memory capacity, we offload them to SSD storage, following common edge deployment practice.
On RTX~4090, we evaluate under constrained GPU memory, where cold experts are offloaded to CPU memory.


\paragraph{Baselines.}
We compare against four baselines:
(i) standard autoregressive decoding (\textbf{AR});
(ii) representative self-speculative decoding with expert sparsity (\textbf{ES}), which constructs a lightweight draft by activating fewer experts per layer than the target model;
(iii) self-speculative decoding with layer sparsity (\textbf{LS}), which forms the draft by skipping a subset of transformer layers selected via Bayesian optimization. Due to its consistently poor efficiency in experiments, LS is evaluated only on DeepSeek and OLMoE, and omitted on Qwen3 and GPT-OSS where it is unlikely to be competitive;
(iv) the state-of-the-art speculative decoding method EAGLE-3 (\textbf{EAGLE3});
(v) a representative MoE-aware speculative decoding approach (\textbf{Cascade}), which selectively enables speculative decoding during inference to balance draft benefit and verification overhead. For a stronger and more competitive baseline, we instantiate Cascade on top of EAGLE3 as the underlying speculative decoding engine.

\subsection{Main Results}

\subsubsection{Efficiency}

\begin{figure*}[!t]
    \vspace{-4pt}
    \centering
    \begin{tabular}{ccccc}
    
        \textcolor{ES}{\rule{1em}{1em}} ES & 
        \textcolor{LS}{\rule{1em}{1em}} LS & 
        \textcolor{EAGLE3}{\rule{1em}{1em}} EAGLE3 & 
        \textcolor{Cascade}{\rule{1em}{1em}} Cascade &
        \textcolor{S2-MoE}{\rule{1em}{1em}} \method\\
    \end{tabular} \\[0.5em]
    \includegraphics[width=2\columnwidth]{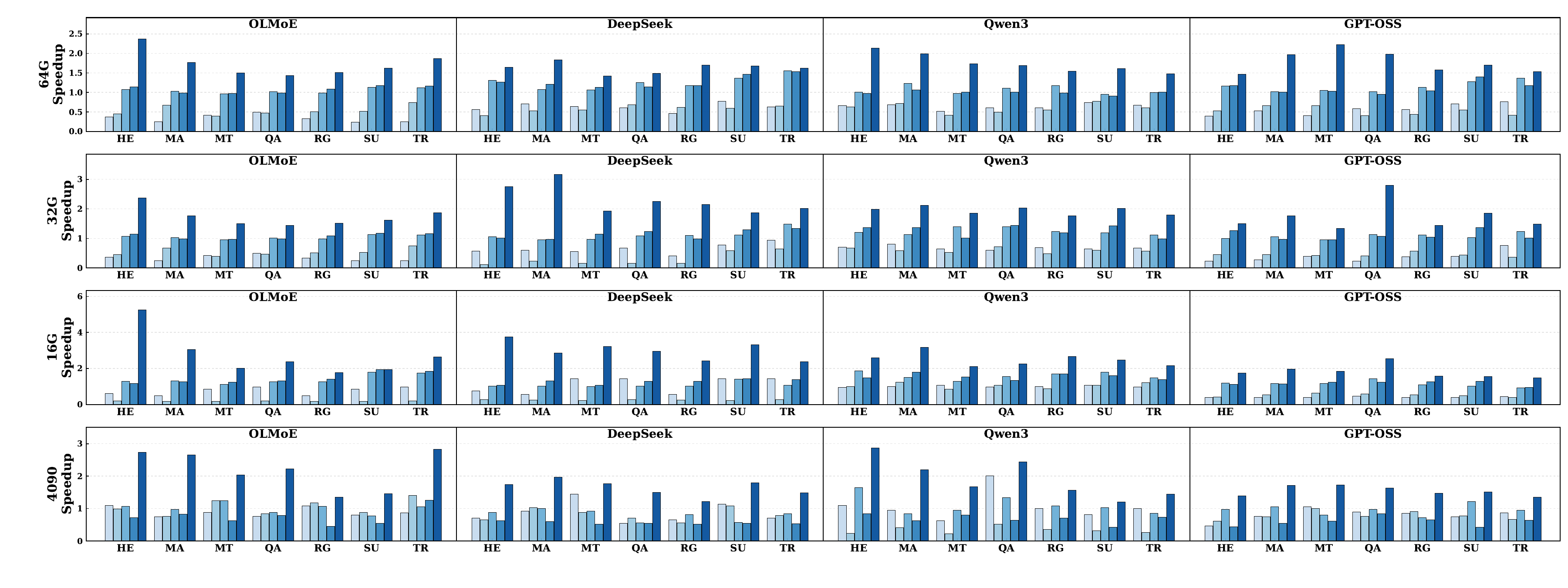}
    \vspace{-8pt}
    \caption{End-to-end speedup of different SD methods across MoE models, tasks, memory budgets on Jetson (16G/32G/64G), and RTX~4090.}
    \label{fig:e2e}
    \vspace{-0.2cm}
\end{figure*}

Fig.~\ref{fig:e2e} summarizes the effectiveness of \method\ across different MoE models, tasks, and memory constraints. 
Since our goal is to improve end-to-end inference efficiency under memory-constrained edge deployment, we use \textbf{Speedup Ratio} relative to standard autoregressive decoding as the metric. Overall, \method\ consistently outperforms autoregressive decoding and other SD baselines, achieving $1.3\text{--}5.3\times$ speedup on Jetson Orin and $1.2\text{--}2.9\times$ on RTX~4090.
Corresponding raw Tok/s, acceptance lengths, and speedups are reported in Appendix~\ref{app:raw_data}.

Naïve self-speculative baselines often fail to outperform autoregressive decoding,
particularly under tight memory budgets.
Although self-speculative decoding improves alignment between draft and target,
its benefits are largely offset by low expert reuse and excessive redundant verification,
leaving substantial optimization headroom.
In contrast, \method\ explicitly targets these bottlenecks, translating aligned drafts into
effective speedups through reduced verification cost and improved parameter reuse.

Compared to EAGLE-3, \method\ achieves more stable gains across a diverse set of modern MoE models.
While EAGLE-3 performs well on some model families (e.g., the Llama series), its effectiveness varies significantly with draft prediction quality and training characteristics.
Despite being training-free, \method\ achieves consistently higher speedups across models and tasks.

Compared to Cascade, which decides whether to enable speculative decoding based on historical information, \method\ performs token-level speculative control using routing-aware cost estimation at the current step. This avoids the extra online decision overhead of test-and-set policies and provides a more direct estimate of verification utility, leading to more consistent speedup gains across models and tasks.

These advantages become more pronounced under tighter memory budgets, where more experts must be offloaded, and verification becomes increasingly sensitive to redundant expert activation and parameter reuse.

\subsubsection{Quality}

Reuse-aware expert gating intentionally introduces a bounded routing bias to improve expert reuse, and is therefore a controlled target-routing approximation rather than a strictly lossless transformation.
Table~\ref{tab:accuracy_all} evaluates its quality impact against the unmodified Original model on both task accuracy and long-context / distribution-consistency metrics.
On standard benchmarks, DeepSeek and OLMoE are evaluated on GSM8K, HumanEval, ARC-E, and ARC-C, while Qwen3 and GPT-OSS are evaluated on AIME24, AIME25, HMMT, and GPQA.
We further evaluate long-context quality on three LongBench~\citep{bai2024longbench} datasets---GovReport, NarrativeQA, and Qasper---and measure output-distribution consistency against the unmodified target on WikiText-2~\citep{merity2017pointer} via Mean KL, RMS logit shift ($\Delta p$), Top-1 agreement, and PPL ratio.
Across models, task-accuracy differences remain small and non-systematic, and LongBench scores stay comparable to the original model.
For the measured models, the PPL ratio remains close to $1.0$ ($1.012$--$1.013$) and Top-1 agreement remains high ($>89\%$), with only small KL/logit shifts, indicating bounded output-level perturbation without observable degradation.

For GPT-OSS, we omit PPL/KL because standard next-token likelihood on raw or non-Harmony continuations is not a reliable quality signal: GPT-OSS is trained for the Harmony chat format and CoT/RL post-training objectives rather than raw language modeling~\citep{openai2025gptoss120bgptoss20bmodel}.
Public reports document anomalously high raw-corpus PPL and poorly calibrated token log-probabilities for GPT-OSS under standard causal-LM evaluation~\citep{hf_transformers_issue_40990,llamacpp_issue_15155}, and recent analysis treats this as a known artifact when Harmony-trained models are evaluated outside their intended format~\citep{zagitov2026tensor}.

\begin{table*}[t]
\centering
\caption{Quality comparison across MoE models: task accuracy, long-context quality, and distribution consistency.}
\label{tab:accuracy_all}
\scriptsize
\setlength{\tabcolsep}{2.2pt}
\renewcommand{\arraystretch}{0.95}
\resizebox{\textwidth}{!}{%
\begin{tabular}{llcccccccccccc}
\toprule
& & \multicolumn{5}{c}{Task accuracy} & \multicolumn{3}{c}{Long-context} & \multicolumn{4}{c}{Distribution consistency} \\
\cmidrule(lr){3-7} \cmidrule(lr){8-10} \cmidrule(lr){11-14}
Model & Method & GSM8K & HumanEval & ARC-E & ARC-C & Avg. & Gov. & NQA & Qasper & Mean KL~$\downarrow$ & RMS~$\Delta p$~$\downarrow$ & Top-1~$\uparrow$ & PPL~$\downarrow$ \\
\midrule
OLMoE
  & Original & 35.29 & 22.56 & 55.81 & 43.86 & 39.38 & 24.69 & 8.10 & 21.52 & 0 & 0 & 100 & 1.000 \\
  & \method  & 34.49 & 24.39 & 55.71 & 43.86 & \textbf{39.61} & 24.14 & 7.90 & 23.30 & 0.048 & 5.20 & 89.89 & 1.012 \\
\midrule
DeepSeek
  & Original & 66.64 & 48.98 & 63.06 & 56.03 & 58.68 & 30.30 & 11.86 & 27.40 & 0 & 0 & 100 & 1.000 \\
  & \method  & 67.27 & 50.20 & 66.91 & 58.56 & \textbf{60.74} & 29.20 & 13.13 & 32.72 & 0.065 & 6.95 & 89.13 & 1.013 \\
\bottomrule
\end{tabular}}

\vspace{6pt}

\resizebox{\textwidth}{!}{%
\begin{tabular}{llcccccccccccc}
\toprule
& & \multicolumn{5}{c}{Task accuracy} & \multicolumn{3}{c}{Long-context} & \multicolumn{4}{c}{Distribution consistency} \\
\cmidrule(lr){3-7} \cmidrule(lr){8-10} \cmidrule(lr){11-14}
Model & Method & AIME24 & AIME25 & HMMT & GPQA & Avg. & Gov. & NQA & Qasper & Mean KL~$\downarrow$ & RMS~$\Delta p$~$\downarrow$ & Top-1~$\uparrow$ & PPL~$\downarrow$ \\
\midrule
Qwen3
  & Original & 68.89 & 54.44 & 36.67 & 79.04 & 59.76 & 30.38 & 15.37 & 28.05 & 0 & 0 & 100 & 1.000 \\
  & \method  & 70.00 & 55.56 & 38.89 & 78.54 & \textbf{60.75} & 30.22 & 16.06 & 27.53 & 0.026 & 4.88 & 93.04 & 1.012 \\
\midrule
GPT-OSS
  & Original & 61.11 & 62.22 & 55.83 & 72.02 & 62.80 & 23.42 & 13.34 & 31.36 & -- & -- & -- & -- \\
  & \method  & 63.34 & 62.22 & 55.00 & 72.01 & \textbf{63.14} & 25.32 & 16.71 & 35.82 & -- & -- & -- & -- \\
\bottomrule
\end{tabular}}
\end{table*}

\subsection{Ablation Study}

\subsubsection{Progressive Speedup Breakdown}

\begin{figure}[h]
    \vspace{-4pt}
    \centering
    \includegraphics[width=\columnwidth]{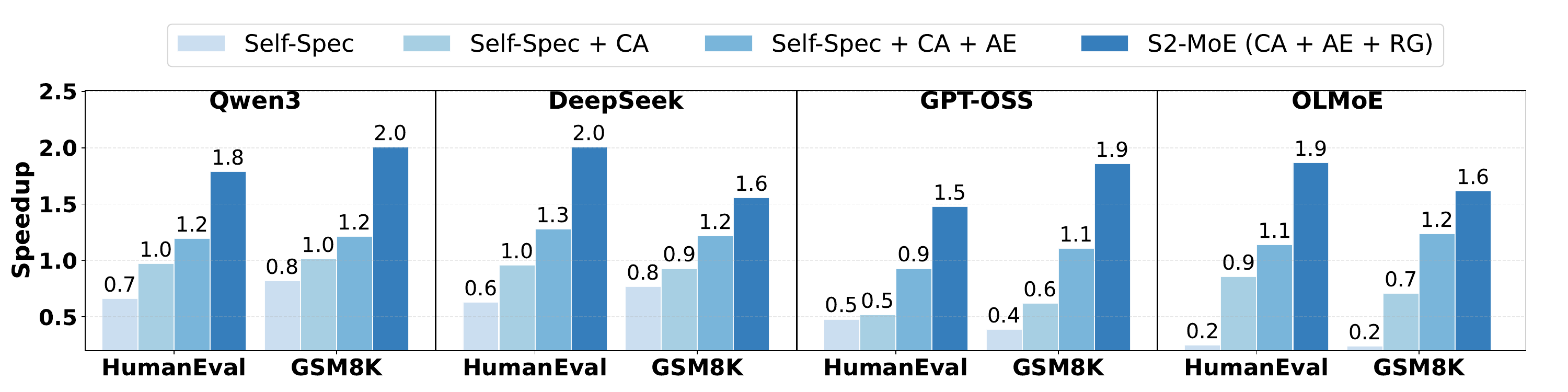}
    \vspace{-8pt}
    \caption{Ablation study on different models, showing the progressive contribution
of each component in \method. CA: context-aligned self-speculative decoding;
AE: adaptive expansion; RG: reuse-aware gating.}
    \label{fig:ablation}
    \vspace{-4pt}
\end{figure}

Figure~\ref{fig:ablation} visualizes the progressive speedup gains contributed by each
component of \method.
Across both Qwen3 and DeepSeek, each design choice consistently improves performance,
and their combination yields the highest overall speedup, confirming that the three
components are complementary rather than redundant.

\subsubsection{Effect of Individual Design Components}

\paragraph{Higher Draft Fidelity.}

We first examine the effect of context-aligned self-speculative decoding on draft fidelity.
Table~\ref{tab:ablation_context_aligned} compares acceptance rates with and without
context alignment.
Across both DeepSeek and OLMoE, enabling context alignment consistently improves acceptance rates on all evaluated tasks.
The improvement is particularly pronounced on reasoning-heavy benchmarks such as GSM8K,
where acceptance rates increase by up to 79\%.
These results indicate that sharing an identical KV cache between the draft and target
effectively prevents error accumulation across speculative iterations, forming a critical foundation for efficient speculative decoding under lightweight drafts.
\begin{table}[H]
\centering
\caption{Effect of context-aligned (CA) self-speculative decoding on acceptance rate (Acc.).
Higher values indicate better draft fidelity.}
\label{tab:ablation_context_aligned}
\footnotesize
\setlength{\tabcolsep}{2pt}
\begin{tabular}{llccc}
\toprule
Model & Dataset & w/o Context-Aligned & w/ Context-Aligned & Acceptance rate $\uparrow$ \\
\midrule
DeepSeek & HumanEval & 43.01 & 52.91 & +23.02\% \\
 & GSM8K     & 36.08 & 48.71 & +35.01\% \\
\midrule
OLMoE    & HumanEval & 32.46 & 40.31 & +24.18\% \\
  & GSM8K     & 33.01 & 59.09 & +79.01\% \\
\bottomrule
\end{tabular}
\vspace{-10pt}
\end{table}
\begin{figure}[h]
    \vspace{-4pt}
    \centering
    \includegraphics[width=\columnwidth]{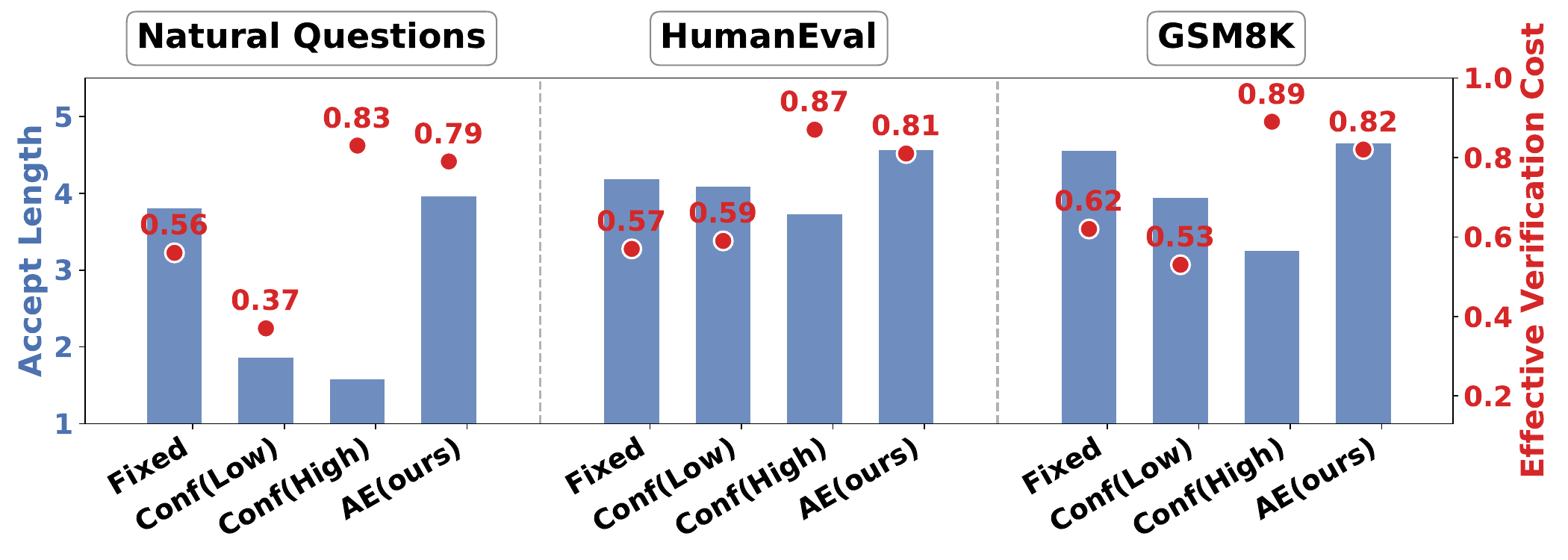}
    \vspace{-6pt}
    \caption{Comparison of different speculative expansion strategies across datasets.
We report the acceptance length and the effective verification cost,
defined as the fraction of expert activations contributing to accepted tokens.}
    \label{fig:prune}
    \vspace{-10pt}
\end{figure}

\paragraph{Less Redundant Verification.}

Figure~\ref{fig:prune} compares different speculative expansion strategies in terms of
effective acceptance length and verification efficiency across multiple datasets.
Fixed-width expansion achieves reasonable acceptance length but incurs substantial
redundant verification.
Confidence-based pruning exhibits a clear trade-off: a low threshold preserves
acceptance but incurs heavy redundant verification, while a higher one
reduces verification overhead but significantly limits acceptance length.

In contrast, our utility-guided adaptive expansion consistently achieves acceptance
length comparable to or exceeding fixed-width expansion, while maintaining verification
efficiency close to aggressive confidence pruning. Compared to confidence-based pruning, our method attains a strictly better
acceptance–verification trade-off, effectively pushing the Pareto frontier by jointly considering acceptance likelihood and verification cost.

\paragraph{Higher Expert Reuse.}

Table~\ref{tab:expert_reuse} reports the effect of our reuse-aware expert gating on expert parameter reuse in DeepSeek.
Across different datasets, enabling reuse-aware gating consistently reduces the reuse ratio, indicating substantially higher expert reuse during verification.
This directly validates Observation~3 in Sec.~\ref{sec:motivation} and confirms that reuse-aware gating effectively aligns expert selection across speculative tokens, thereby improving parameter locality.


\subsubsection{Hyperparameter Sensitivity Analysis}

We further study the sensitivity of two key hyperparameters in \method: the draft expert top-$k$ and the reuse-aware gating cap.

\begin{figure}[h]
    \vspace{-4pt}
    \centering
    \includegraphics[width=\columnwidth]{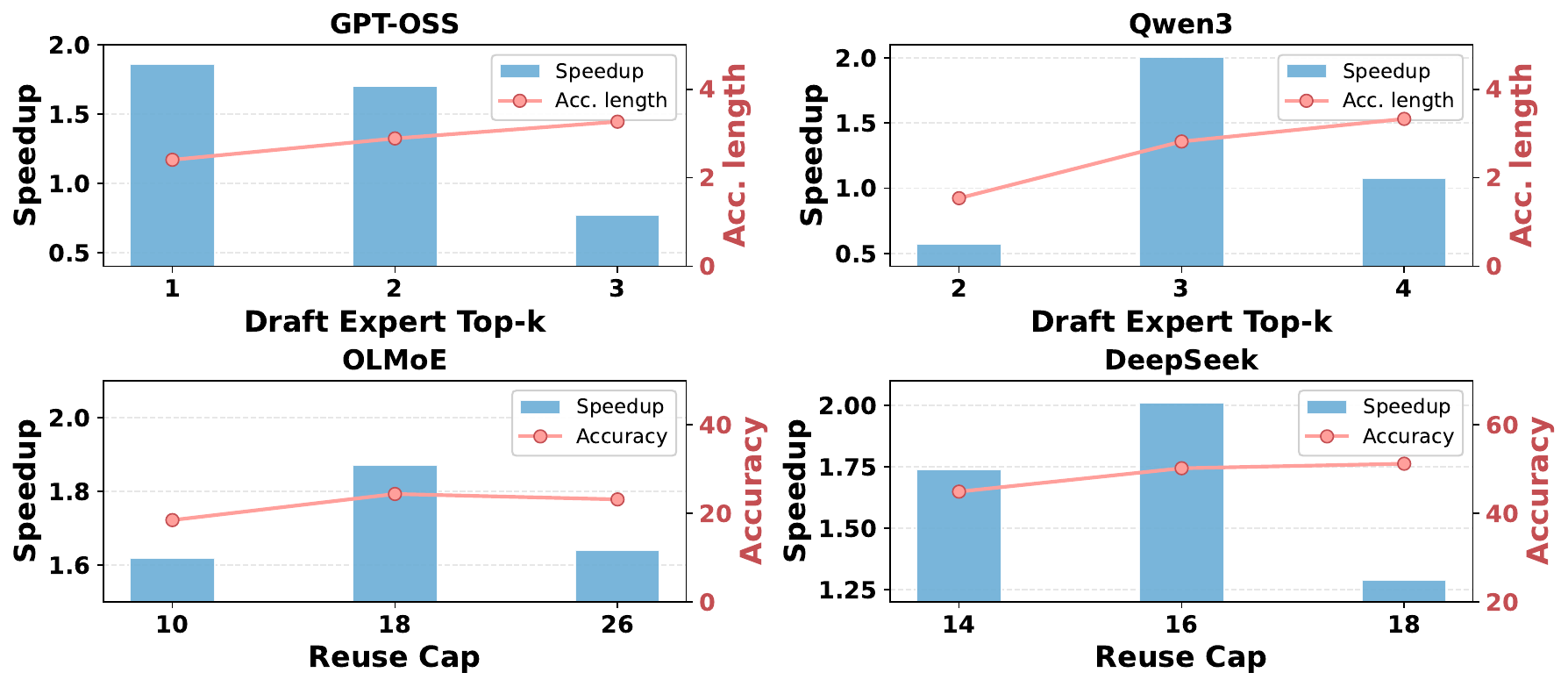}
    \vspace{-6pt}
    \caption{Sensitivity to key hyperparameters in \method.}
    \label{fig:para}
    \vspace{-10pt}
\end{figure}

\paragraph{Effect of draft expert top-$k$.}
The top row of Fig.~\ref{fig:para} studies the effect of draft expert top-$k$, i.e., the number of experts activated per token in the self-speculative draft model, on Qwen3 and GPT-OSS on GSM8K.
Increasing top-$k$ improves acceptance length by providing higher-quality draft tokens, but also increases draft overhead,
which can reduce overall speedup.
Conversely, overly small top-$k$ limits acceptance length and underutilizes speculative opportunities.
These results indicate a clear trade-off, and motivate choosing a moderate
top-$k$ that balances draft cost and acceptance gain.

\begin{table}[H]
\centering
\small
\caption{Effect of reuse-aware expert gating on expert parameter reuse.
Expert reuse is measured using the reuse ratio defined in
Challenge~2 (Sec.~\ref{sec:motivation}).}
\label{tab:expert_reuse}

\setlength{\tabcolsep}{5pt}
\begin{tabular}{l c c c}
\toprule
Dataset & w/o RG & w/ RG & Reuse Improvement (\%) \\
\midrule
HumanEval & 0.6017 & 0.4453 & 25.99 \\
 GSM8K     & 0.5893 & 0.3943 & 33.09 \\
\bottomrule
\end{tabular}
\end{table}

\paragraph{Effect of reuse-aware gating $\texttt{cap}$.}
The bottom row of Fig.~\ref{fig:para} evaluates the impact of the reuse-aware gating $\texttt{cap}$ on DeepSeek and OLMoE on HumanEval.
When the $\texttt{cap}$ is too small, only a limited number of experts are encouraged for reuse, while the remaining experts are still selected in a scattered manner, leading to insufficient improvement in expert reuse.
On the other hand, an excessively large $\texttt{cap}$ dilutes the bias by rewarding too many experts, effectively weakening the reuse signal and reducing its impact on verification efficiency.
In practice, we find that setting the $\texttt{cap}$ within a moderate range (e.g., $1.5\times$ to $2.5\times$ the top-$k$) provides a good balance between promoting expert reuse and preserving routing flexibility.

\section{Conclusion}
\label{sec:conclusion}


In conclusion, we present \method, an efficient self-speculative decoding framework for Mixture-of-Experts models in memory-constrained edge environments.
By alleviating the memory-bandwidth bottleneck through context alignment, utility-guided speculation, and reuse-aware gating, \method\ achieves \textbf{1.3$\times$ to 5.3$\times$} speedup on Jetson Orin and \textbf{1.2$\times$ to 2.9$\times$} on RTX~4090 over autoregressive decoding, and consistently outperforms state-of-the-art speculative baselines while maintaining comparable accuracy.

\newpage
\appendix
\section{End-to-End Raw Measurements}
\label{app:raw_data}
{\footnotesize This appendix reports the raw throughput (Tok/s), acceptance length (Acc.), and speedup underlying Fig.~\ref{fig:e2e}.
Tables~\ref{tab:raw-16g}--\ref{tab:raw-4090} list these measurements on Jetson Orin NX 16~GB, Jetson AGX Orin 32~GB/64~GB, and RTX~4090.}

\begin{table*}[!t]
\centering
\begin{minipage}[t]{\columnwidth}
\centering
\captionof{table}{Raw measurements on NVIDIA Jetson Orin NX 16~GB.}
\label{tab:raw-16g}
\vspace{-0.35em}
\adjustbox{max width=\linewidth,max height=0.435\textheight,center}{%
\setlength{\tabcolsep}{0.7pt}%
\renewcommand{\arraystretch}{0.72}%
\scriptsize
\begin{tabular}{@{}c c r r r r r r r r r r r r@{}}
\toprule
& & \multicolumn{3}{c}{DeepSeek} & \multicolumn{3}{c}{OLMoE} & \multicolumn{3}{c}{Qwen3} & \multicolumn{3}{c}{GPT-OSS} \\
\cmidrule(lr){3-5}\cmidrule(lr){6-8}\cmidrule(lr){9-11}\cmidrule(lr){12-14}
Task & Method & T/s & Acc & Spd & T/s & Acc & Spd & T/s & Acc & Spd & T/s & Acc & Spd \\
\midrule
\multirow{6}{*}{HE} & Auto & 0.55 & 1.00 & 1.00 & 2.30 & 1.00 & 1.00 & 0.47 & 1.00 & 1.00 & 0.95 & 1.00 & 1.00 \\
 & ES & 0.41 & 4.73 & 0.74 & 1.40 & 3.08 & 0.61 & 0.45 & 5.00 & 0.94 & 0.38 & 2.77 & 0.40 \\
 & LS & 0.15 & 1.70 & 0.27 & 0.46 & 1.06 & 0.20 & 0.47 & 5.08 & 0.99 & 0.41 & 1.58 & 0.42 \\
 & EAGLE3 & 0.56 & 2.43 & 1.02 & 2.96 & 1.23 & 1.29 & 0.89 & 1.15 & 1.87 & 1.14 & 1.81 & 1.19 \\
 & Cascade & 0.58 & 1.10 & 1.07 & 2.69 & 1.26 & 1.17 & 0.70 & 1.96 & 1.48 & 1.07 & 1.00 & 1.12 \\
 & \method & 2.07 & 5.42 & 3.76 & 12.08 & 7.00 & 5.25 & 1.22 & 6.70 & 2.60 & 1.65 & 2.95 & 1.74 \\
\multirow{6}{*}{MA} & Auto & 0.69 & 1.00 & 1.00 & 2.22 & 1.00 & 1.00 & 0.41 & 1.00 & 1.00 & 0.99 & 1.00 & 1.00 \\
 & ES & 0.39 & 4.36 & 0.56 & 1.06 & 3.15 & 0.48 & 0.41 & 4.71 & 1.00 & 0.38 & 1.63 & 0.38 \\
 & LS & 0.17 & 1.56 & 0.24 & 0.38 & 1.05 & 0.17 & 0.51 & 3.38 & 1.23 & 0.54 & 2.42 & 0.55 \\
 & EAGLE3 & 0.70 & 1.54 & 1.01 & 2.91 & 1.29 & 1.31 & 0.61 & 1.21 & 1.49 & 1.15 & 1.21 & 1.16 \\
 & Cascade & 0.90 & 1.13 & 1.30 & 2.77 & 1.33 & 1.25 & 0.74 & 1.66 & 1.80 & 1.12 & 1.37 & 1.13 \\
 & \method & 1.97 & 4.06 & 2.85 & 6.77 & 4.79 & 3.05 & 1.30 & 6.45 & 3.17 & 1.95 & 3.25 & 1.96 \\
\multirow{6}{*}{MT} & Auto & 0.49 & 1.00 & 1.00 & 2.10 & 1.00 & 1.00 & 0.48 & 1.00 & 1.00 & 1.00 & 1.00 & 1.00 \\
 & ES & 0.70 & 3.14 & 1.43 & 1.80 & 2.52 & 0.86 & 0.51 & 3.64 & 1.06 & 0.40 & 2.98 & 0.40 \\
 & LS & 0.10 & 1.22 & 0.21 & 0.38 & 4.08 & 0.18 & 0.41 & 2.58 & 0.85 & 0.63 & 2.24 & 0.62 \\
 & EAGLE3 & 0.48 & 1.75 & 0.99 & 2.33 & 1.19 & 1.11 & 0.62 & 1.15 & 1.28 & 1.17 & 1.81 & 1.17 \\
 & Cascade & 0.52 & 1.77 & 1.06 & 2.60 & 1.30 & 1.24 & 0.74 & 1.89 & 1.53 & 1.23 & 1.18 & 1.23 \\
 & \method & 1.57 & 2.79 & 3.21 & 4.22 & 3.35 & 2.01 & 1.01 & 4.64 & 2.10 & 1.85 & 3.05 & 1.85 \\
\multirow{6}{*}{QA} & Auto & 0.61 & 1.00 & 1.00 & 2.09 & 1.00 & 1.00 & 0.50 & 1.00 & 1.00 & 0.97 & 1.00 & 1.00 \\
 & ES & 0.88 & 2.67 & 1.44 & 2.05 & 1.89 & 0.98 & 0.49 & 3.80 & 0.98 & 0.45 & 1.56 & 0.46 \\
 & LS & 0.16 & 1.88 & 0.27 & 0.42 & 4.72 & 0.20 & 0.53 & 3.64 & 1.07 & 0.56 & 1.50 & 0.58 \\
 & EAGLE3 & 0.61 & 1.77 & 1.01 & 2.61 & 1.10 & 1.25 & 0.78 & 1.81 & 1.56 & 1.38 & 1.31 & 1.42 \\
 & Cascade & 0.79 & 1.58 & 1.29 & 2.74 & 1.20 & 1.31 & 0.67 & 1.70 & 1.34 & 1.20 & 1.24 & 1.23 \\
 & \method & 1.80 & 3.88 & 2.94 & 4.95 & 3.94 & 2.37 & 1.13 & 5.15 & 2.26 & 2.45 & 3.35 & 2.53 \\
\multirow{6}{*}{RG} & Auto & 0.58 & 1.00 & 1.00 & 2.12 & 1.00 & 1.00 & 0.49 & 1.00 & 1.00 & 0.98 & 1.00 & 1.00 \\
 & ES & 0.33 & 2.67 & 0.56 & 1.02 & 1.89 & 0.48 & 0.49 & 3.80 & 1.00 & 0.37 & 1.56 & 0.38 \\
 & LS & 0.14 & 1.88 & 0.24 & 0.36 & 2.00 & 0.17 & 0.43 & 2.78 & 0.88 & 0.52 & 1.28 & 0.53 \\
 & EAGLE3 & 0.60 & 1.91 & 1.03 & 2.69 & 1.09 & 1.27 & 0.82 & 1.07 & 1.69 & 1.05 & 1.09 & 1.08 \\
 & Cascade & 0.75 & 1.63 & 1.28 & 2.96 & 1.16 & 1.40 & 0.83 & 1.81 & 1.70 & 1.22 & 1.13 & 1.25 \\
 & \method & 1.40 & 4.46 & 2.42 & 3.75 & 5.00 & 1.77 & 1.31 & 5.38 & 2.67 & 1.54 & 2.71 & 1.57 \\
\multirow{6}{*}{SU} & Auto & 0.47 & 1.00 & 1.00 & 2.11 & 1.00 & 1.00 & 0.48 & 1.00 & 1.00 & 0.98 & 1.00 & 1.00 \\
 & ES & 0.68 & 4.36 & 1.43 & 1.81 & 3.15 & 0.86 & 0.51 & 4.71 & 1.06 & 0.39 & 1.63 & 0.40 \\
 & LS & 0.10 & 1.56 & 0.21 & 0.38 & 1.05 & 0.18 & 0.51 & 3.06 & 1.05 & 0.47 & 1.06 & 0.48 \\
 & EAGLE3 & 0.67 & 1.74 & 1.41 & 3.77 & 1.09 & 1.79 & 0.86 & 1.82 & 1.79 & 0.99 & 1.81 & 1.01 \\
 & Cascade & 0.68 & 1.53 & 1.44 & 4.07 & 1.07 & 1.93 & 0.77 & 1.82 & 1.61 & 1.26 & 1.83 & 1.28 \\
 & \method & 1.56 & 3.72 & 3.31 & 4.09 & 3.82 & 1.94 & 1.18 & 5.15 & 2.46 & 1.53 & 2.39 & 1.56 \\
\multirow{6}{*}{TR} & Auto & 0.61 & 1.00 & 1.00 & 2.14 & 1.00 & 1.00 & 0.48 & 1.00 & 1.00 & 0.98 & 1.00 & 1.00 \\
 & ES & 0.88 & 3.14 & 1.44 & 2.10 & 2.52 & 0.98 & 0.47 & 3.64 & 0.98 & 0.42 & 2.98 & 0.43 \\
 & LS & 0.17 & 1.22 & 0.27 & 0.43 & 1.70 & 0.20 & 0.58 & 3.87 & 1.20 & 0.39 & 1.51 & 0.40 \\
 & EAGLE3 & 0.66 & 1.46 & 1.07 & 3.75 & 1.09 & 1.75 & 0.70 & 1.06 & 1.47 & 0.90 & 1.75 & 0.91 \\
 & Cascade & 0.85 & 1.52 & 1.38 & 3.96 & 1.11 & 1.85 & 0.66 & 1.82 & 1.38 & 0.94 & 1.64 & 0.95 \\
 & \method & 1.45 & 3.82 & 2.38 & 5.65 & 3.63 & 2.64 & 1.04 & 6.27 & 2.16 & 1.44 & 2.39 & 1.47 \\
\bottomrule
\end{tabular}%
}
\vspace{0.35em}
\captionof{table}{Raw measurements on NVIDIA Jetson AGX Orin 32~GB.}
\label{tab:raw-32g}
\vspace{-0.35em}
\adjustbox{max width=\linewidth,max height=0.435\textheight,center}{%
\setlength{\tabcolsep}{0.7pt}%
\renewcommand{\arraystretch}{0.72}%
\scriptsize
\begin{tabular}{@{}c c r r r r r r r r r r r r@{}}
\toprule
& & \multicolumn{3}{c}{DeepSeek} & \multicolumn{3}{c}{OLMoE} & \multicolumn{3}{c}{Qwen3} & \multicolumn{3}{c}{GPT-OSS} \\
\cmidrule(lr){3-5}\cmidrule(lr){6-8}\cmidrule(lr){9-11}\cmidrule(lr){12-14}
Task & Method & T/s & Acc & Spd & T/s & Acc & Spd & T/s & Acc & Spd & T/s & Acc & Spd \\
\midrule
\multirow{6}{*}{HE} & Auto & 2.30 & 1.00 & 1.00 & 31.27 & 1.00 & 1.00 & 0.78 & 1.00 & 1.00 & 1.12 & 1.00 & 1.00 \\
 & ES & 1.31 & 4.62 & 0.57 & 11.57 & 1.03 & 0.37 & 0.55 & 4.14 & 0.70 & 0.26 & 2.54 & 0.23 \\
 & LS & 0.28 & 6.92 & 0.12 & 14.07 & 6.55 & 0.45 & 0.53 & 4.46 & 0.68 & 0.51 & 1.96 & 0.45 \\
 & EAGLE3 & 2.44 & 2.46 & 1.06 & 33.46 & 1.27 & 1.07 & 0.95 & 1.50 & 1.21 & 1.12 & 2.36 & 1.00 \\
 & Cascade & 2.32 & 1.96 & 1.01 & 35.65 & 1.31 & 1.14 & 1.07 & 1.02 & 1.36 & 1.43 & 2.43 & 1.27 \\
 & \method & 6.32 & 7.78 & 2.75 & 74.32 & 5.23 & 2.38 & 1.54 & 5.00 & 1.98 & 1.69 & 4.47 & 1.50 \\
\multirow{6}{*}{MA} & Auto & 2.38 & 1.00 & 1.00 & 31.34 & 1.00 & 1.00 & 0.79 & 1.00 & 1.00 & 1.13 & 1.00 & 1.00 \\
 & ES & 1.43 & 3.91 & 0.60 & 7.83 & 1.07 & 0.25 & 0.63 & 3.19 & 0.80 & 0.31 & 1.88 & 0.27 \\
 & LS & 0.55 & 6.37 & 0.23 & 21.00 & 4.63 & 0.67 & 0.47 & 3.25 & 0.59 & 0.51 & 1.87 & 0.45 \\
 & EAGLE3 & 2.27 & 1.87 & 0.95 & 32.28 & 1.35 & 1.03 & 0.89 & 1.95 & 1.13 & 1.19 & 1.89 & 1.05 \\
 & Cascade & 2.31 & 1.79 & 0.97 & 30.71 & 1.26 & 0.98 & 1.07 & 1.94 & 1.36 & 1.10 & 1.16 & 0.97 \\
 & \method & 7.53 & 6.25 & 3.16 & 55.37 & 4.47 & 1.77 & 1.67 & 4.38 & 2.12 & 1.99 & 5.42 & 1.76 \\
\multirow{6}{*}{MT} & Auto & 2.24 & 1.00 & 1.00 & 31.28 & 1.00 & 1.00 & 0.78 & 1.00 & 1.00 & 1.15 & 1.00 & 1.00 \\
 & ES & 1.23 & 3.16 & 0.55 & 13.14 & 2.24 & 0.42 & 0.51 & 3.69 & 0.65 & 0.45 & 1.07 & 0.39 \\
 & LS & 0.34 & 1.18 & 0.15 & 12.20 & 4.08 & 0.39 & 0.41 & 2.35 & 0.52 & 0.49 & 2.47 & 0.43 \\
 & EAGLE3 & 2.18 & 1.91 & 0.97 & 30.03 & 1.15 & 0.96 & 1.08 & 1.02 & 1.39 & 1.09 & 1.22 & 0.95 \\
 & Cascade & 2.56 & 1.76 & 1.14 & 30.34 & 1.28 & 0.97 & 0.79 & 1.83 & 1.01 & 1.10 & 1.13 & 0.96 \\
 & \method & 4.31 & 5.29 & 1.92 & 46.94 & 2.95 & 1.50 & 1.45 & 4.47 & 1.86 & 1.54 & 3.61 & 1.34 \\
\multirow{6}{*}{QA} & Auto & 2.19 & 1.00 & 1.00 & 31.30 & 1.00 & 1.00 & 0.76 & 1.00 & 1.00 & 1.10 & 1.00 & 1.00 \\
 & ES & 1.47 & 2.09 & 0.67 & 15.34 & 1.92 & 0.49 & 0.45 & 3.49 & 0.60 & 0.25 & 2.21 & 0.23 \\
 & LS & 0.33 & 1.54 & 0.15 & 14.71 & 4.72 & 0.47 & 0.55 & 3.47 & 0.72 & 0.44 & 1.24 & 0.40 \\
 & EAGLE3 & 2.39 & 1.81 & 1.09 & 31.93 & 1.20 & 1.02 & 1.06 & 1.84 & 1.40 & 1.24 & 1.12 & 1.13 \\
 & Cascade & 2.69 & 1.54 & 1.23 & 30.99 & 1.21 & 0.99 & 1.09 & 1.57 & 1.44 & 1.17 & 1.26 & 1.07 \\
 & \method & 4.95 & 4.56 & 2.26 & 44.92 & 3.50 & 1.44 & 1.54 & 4.12 & 2.03 & 3.08 & 4.60 & 2.80 \\
\multirow{6}{*}{RG} & Auto & 2.20 & 1.00 & 1.00 & 31.27 & 1.00 & 1.00 & 0.78 & 1.00 & 1.00 & 1.12 & 1.00 & 1.00 \\
 & ES & 0.90 & 2.09 & 0.41 & 10.32 & 1.92 & 0.33 & 0.54 & 3.49 & 0.69 & 0.43 & 2.21 & 0.38 \\
 & LS & 0.33 & 1.54 & 0.15 & 15.95 & 4.72 & 0.51 & 0.37 & 2.19 & 0.47 & 0.64 & 1.73 & 0.57 \\
 & EAGLE3 & 2.42 & 1.96 & 1.10 & 30.96 & 1.17 & 0.99 & 0.96 & 1.99 & 1.23 & 1.24 & 1.18 & 1.11 \\
 & Cascade & 2.18 & 1.61 & 0.99 & 34.09 & 1.15 & 1.09 & 0.93 & 1.90 & 1.19 & 1.17 & 1.14 & 1.04 \\
 & \method & 4.71 & 5.91 & 2.14 & 47.20 & 3.67 & 1.51 & 1.38 & 4.31 & 1.77 & 1.61 & 3.14 & 1.44 \\
\multirow{6}{*}{SU} & Auto & 2.25 & 1.00 & 1.00 & 31.37 & 1.00 & 1.00 & 0.79 & 1.00 & 1.00 & 1.14 & 1.00 & 1.00 \\
 & ES & 1.74 & 3.91 & 0.77 & 7.53 & 1.07 & 0.24 & 0.51 & 3.19 & 0.65 & 0.45 & 1.88 & 0.39 \\
 & LS & 1.33 & 6.37 & 0.59 & 16.31 & 4.63 & 0.52 & 0.47 & 3.33 & 0.59 & 0.49 & 1.83 & 0.43 \\
 & EAGLE3 & 2.50 & 1.61 & 1.11 & 35.45 & 1.07 & 1.13 & 0.94 & 1.83 & 1.19 & 1.18 & 1.23 & 1.03 \\
 & Cascade & 2.93 & 1.71 & 1.30 & 37.02 & 1.12 & 1.18 & 1.12 & 1.67 & 1.42 & 1.55 & 1.24 & 1.36 \\
 & \method & 4.21 & 3.24 & 1.87 & 50.82 & 3.05 & 1.62 & 1.59 & 3.89 & 2.01 & 2.12 & 3.00 & 1.86 \\
\multirow{6}{*}{TR} & Auto & 2.26 & 1.00 & 1.00 & 31.32 & 1.00 & 1.00 & 0.78 & 1.00 & 1.00 & 1.10 & 1.00 & 1.00 \\
 & ES & 2.12 & 3.16 & 0.94 & 7.83 & 2.24 & 0.25 & 0.52 & 3.69 & 0.67 & 0.84 & 1.07 & 0.76 \\
 & LS & 1.47 & 1.18 & 0.65 & 23.18 & 4.08 & 0.74 & 0.44 & 3.60 & 0.57 & 0.41 & 1.16 & 0.37 \\
 & EAGLE3 & 3.34 & 1.55 & 1.48 & 35.08 & 1.04 & 1.12 & 0.86 & 1.30 & 1.11 & 1.37 & 1.81 & 1.24 \\
 & Cascade & 3.00 & 1.46 & 1.33 & 36.34 & 1.22 & 1.16 & 0.77 & 1.90 & 0.99 & 1.12 & 1.73 & 1.02 \\
 & \method & 4.54 & 4.17 & 2.01 & 58.57 & 2.88 & 1.87 & 1.40 & 5.31 & 1.79 & 1.63 & 4.12 & 1.48 \\
\bottomrule
\end{tabular}%
}
\end{minipage}
\hfill
\begin{minipage}[t]{\columnwidth}
\centering
\captionof{table}{Raw measurements on NVIDIA Jetson AGX Orin 64~GB.}
\label{tab:raw-64g}
\vspace{-0.35em}
\adjustbox{max width=\linewidth,max height=0.435\textheight,center}{%
\setlength{\tabcolsep}{0.7pt}%
\renewcommand{\arraystretch}{0.72}%
\scriptsize
\begin{tabular}{@{}c c r r r r r r r r r r r r@{}}
\toprule
& & \multicolumn{3}{c}{DeepSeek} & \multicolumn{3}{c}{OLMoE} & \multicolumn{3}{c}{Qwen3} & \multicolumn{3}{c}{GPT-OSS} \\
\cmidrule(lr){3-5}\cmidrule(lr){6-8}\cmidrule(lr){9-11}\cmidrule(lr){12-14}
Task & Method & T/s & Acc & Spd & T/s & Acc & Spd & T/s & Acc & Spd & T/s & Acc & Spd \\
\midrule
\multirow{6}{*}{HE} & Auto & 16.17 & 1.00 & 1.00 & 31.36 & 1.00 & 1.00 & 0.85 & 1.00 & 1.00 & 1.30 & 1.00 & 1.00 \\
 & ES & 9.06 & 8.00 & 0.56 & 11.60 & 1.03 & 0.37 & 0.56 & 6.67 & 0.66 & 0.51 & 2.33 & 0.39 \\
 & LS & 6.63 & 8.00 & 0.41 & 14.11 & 6.55 & 0.45 & 0.54 & 3.71 & 0.63 & 0.69 & 2.09 & 0.53 \\
 & EAGLE3 & 21.18 & 2.69 & 1.31 & 33.55 & 1.27 & 1.07 & 0.86 & 1.01 & 1.01 & 1.50 & 1.01 & 1.16 \\
 & Cascade & 20.38 & 1.51 & 1.26 & 35.75 & 1.31 & 1.14 & 0.83 & 1.95 & 0.97 & 1.52 & 1.01 & 1.17 \\
 & \method & 26.62 & 5.50 & 1.65 & 74.32 & 5.23 & 2.37 & 1.82 & 5.00 & 2.14 & 1.91 & 2.83 & 1.47 \\
\multirow{6}{*}{MA} & Auto & 16.18 & 1.00 & 1.00 & 31.28 & 1.00 & 1.00 & 0.90 & 1.00 & 1.00 & 1.30 & 1.00 & 1.00 \\
 & ES & 11.48 & 2.80 & 0.71 & 7.82 & 1.07 & 0.25 & 0.61 & 3.00 & 0.68 & 0.69 & 1.67 & 0.53 \\
 & LS & 8.57 & 1.67 & 0.53 & 20.96 & 4.63 & 0.67 & 0.65 & 3.86 & 0.72 & 0.86 & 2.38 & 0.66 \\
 & EAGLE3 & 17.31 & 1.32 & 1.07 & 32.22 & 1.35 & 1.03 & 1.11 & 1.17 & 1.23 & 1.33 & 1.65 & 1.02 \\
 & Cascade & 19.57 & 1.40 & 1.21 & 30.66 & 1.26 & 0.98 & 0.96 & 1.07 & 1.06 & 1.32 & 1.38 & 1.01 \\
 & \method & 29.68 & 5.00 & 1.83 & 55.37 & 4.47 & 1.77 & 1.79 & 4.38 & 1.99 & 2.56 & 3.25 & 1.97 \\
\multirow{6}{*}{MT} & Auto & 16.18 & 1.00 & 1.00 & 31.31 & 1.00 & 1.00 & 0.91 & 1.00 & 1.00 & 1.05 & 1.00 & 1.00 \\
 & ES & 10.36 & 1.29 & 0.64 & 13.15 & 2.24 & 0.42 & 0.47 & 7.00 & 0.52 & 0.43 & 2.00 & 0.41 \\
 & LS & 8.90 & 1.18 & 0.55 & 12.21 & 4.08 & 0.39 & 0.38 & 1.91 & 0.41 & 0.69 & 1.58 & 0.66 \\
 & EAGLE3 & 17.15 & 1.67 & 1.06 & 30.06 & 1.15 & 0.96 & 0.88 & 1.13 & 0.97 & 1.10 & 1.11 & 1.05 \\
 & Cascade & 18.29 & 1.64 & 1.13 & 30.37 & 1.28 & 0.97 & 0.92 & 1.81 & 1.01 & 1.08 & 1.11 & 1.03 \\
 & \method & 22.91 & 3.63 & 1.42 & 46.94 & 2.95 & 1.50 & 1.58 & 4.47 & 1.74 & 2.33 & 2.83 & 2.22 \\
\multirow{6}{*}{QA} & Auto & 16.16 & 1.00 & 1.00 & 31.29 & 1.00 & 1.00 & 0.89 & 1.00 & 1.00 & 1.33 & 1.00 & 1.00 \\
 & ES & 9.86 & 2.60 & 0.61 & 15.33 & 1.92 & 0.49 & 0.54 & 5.33 & 0.60 & 0.77 & 1.25 & 0.58 \\
 & LS & 10.99 & 2.60 & 0.68 & 14.71 & 4.72 & 0.47 & 0.44 & 2.42 & 0.49 & 0.53 & 1.97 & 0.40 \\
 & EAGLE3 & 20.20 & 1.73 & 1.25 & 31.92 & 1.20 & 1.02 & 0.99 & 1.82 & 1.11 & 1.35 & 1.29 & 1.02 \\
 & Cascade & 18.42 & 1.43 & 1.14 & 30.98 & 1.21 & 0.99 & 0.90 & 1.67 & 1.01 & 1.26 & 1.13 & 0.95 \\
 & \method & 24.08 & 3.61 & 1.49 & 44.92 & 3.50 & 1.44 & 1.51 & 4.12 & 1.69 & 2.63 & 3.05 & 1.98 \\
\multirow{6}{*}{RG} & Auto & 16.16 & 1.00 & 1.00 & 31.30 & 1.00 & 1.00 & 0.91 & 1.00 & 1.00 & 1.31 & 1.00 & 1.00 \\
 & ES & 7.44 & 1.57 & 0.46 & 10.33 & 1.92 & 0.33 & 0.54 & 4.00 & 0.60 & 0.74 & 1.60 & 0.56 \\
 & LS & 10.02 & 3.00 & 0.62 & 15.96 & 2.00 & 0.51 & 0.50 & 2.82 & 0.55 & 0.58 & 1.83 & 0.44 \\
 & EAGLE3 & 19.07 & 1.79 & 1.18 & 30.98 & 1.17 & 0.99 & 1.06 & 1.90 & 1.17 & 1.48 & 1.22 & 1.13 \\
 & Cascade & 18.91 & 1.61 & 1.17 & 34.11 & 1.15 & 1.09 & 0.90 & 1.93 & 0.99 & 1.37 & 1.13 & 1.04 \\
 & \method & 27.40 & 4.38 & 1.70 & 47.20 & 3.67 & 1.51 & 1.40 & 4.31 & 1.54 & 2.06 & 2.71 & 1.57 \\
\multirow{6}{*}{SU} & Auto & 16.17 & 1.00 & 1.00 & 31.28 & 1.00 & 1.00 & 0.90 & 1.00 & 1.00 & 1.30 & 1.00 & 1.00 \\
 & ES & 12.45 & 2.20 & 0.77 & 7.51 & 1.43 & 0.24 & 0.67 & 3.00 & 0.74 & 0.92 & 2.40 & 0.71 \\
 & LS & 9.54 & 1.29 & 0.59 & 16.27 & 3.40 & 0.52 & 0.70 & 4.00 & 0.77 & 0.71 & 1.64 & 0.55 \\
 & EAGLE3 & 21.99 & 1.79 & 1.36 & 35.34 & 1.07 & 1.13 & 0.86 & 1.83 & 0.95 & 1.66 & 1.65 & 1.28 \\
 & Cascade & 23.61 & 1.75 & 1.46 & 36.91 & 1.12 & 1.18 & 0.82 & 1.70 & 0.91 & 1.82 & 1.67 & 1.40 \\
 & \method & 27.17 & 3.45 & 1.68 & 50.67 & 3.05 & 1.62 & 1.45 & 3.89 & 1.61 & 2.21 & 2.75 & 1.70 \\
\multirow{6}{*}{TR} & Auto & 16.18 & 1.00 & 1.00 & 31.30 & 1.00 & 1.00 & 0.89 & 1.00 & 1.00 & 1.27 & 1.00 & 1.00 \\
 & ES & 10.19 & 5.33 & 0.63 & 7.82 & 1.89 & 0.25 & 0.59 & 3.50 & 0.67 & 0.97 & 1.67 & 0.76 \\
 & LS & 10.52 & 1.67 & 0.65 & 23.16 & 1.70 & 0.74 & 0.54 & 3.06 & 0.61 & 0.53 & 1.12 & 0.41 \\
 & EAGLE3 & 25.24 & 1.47 & 1.56 & 35.05 & 1.04 & 1.12 & 0.89 & 1.09 & 1.00 & 1.75 & 1.81 & 1.37 \\
 & Cascade & 24.76 & 1.52 & 1.53 & 36.30 & 1.22 & 1.16 & 0.90 & 1.68 & 1.01 & 1.49 & 1.51 & 1.17 \\
 & \method & 26.21 & 3.94 & 1.62 & 58.53 & 2.88 & 1.87 & 1.32 & 5.31 & 1.48 & 1.94 & 2.83 & 1.53 \\
\bottomrule
\end{tabular}%
}
\vspace{0.35em}
\captionof{table}{Raw measurements on NVIDIA RTX~4090.}
\label{tab:raw-4090}
\vspace{-0.35em}
\adjustbox{max width=\linewidth,max height=0.435\textheight,center}{%
\setlength{\tabcolsep}{0.7pt}%
\renewcommand{\arraystretch}{0.72}%
\scriptsize
\begin{tabular}{@{}c c r r r r r r r r r r r r@{}}
\toprule
& & \multicolumn{3}{c}{DeepSeek} & \multicolumn{3}{c}{OLMoE} & \multicolumn{3}{c}{Qwen3} & \multicolumn{3}{c}{GPT-OSS} \\
\cmidrule(lr){3-5}\cmidrule(lr){6-8}\cmidrule(lr){9-11}\cmidrule(lr){12-14}
Task & Method & T/s & Acc & Spd & T/s & Acc & Spd & T/s & Acc & Spd & T/s & Acc & Spd \\
\midrule
\multirow{6}{*}{HE} & Auto & 3.60 & 1.00 & 1.00 & 3.63 & 1.00 & 1.00 & 1.00 & 1.00 & 1.00 & 1.68 & 1.00 & 1.00 \\
 & ES & 2.56 & 3.00 & 0.71 & 3.95 & 3.14 & 1.09 & 1.09 & 4.75 & 1.09 & 0.79 & 1.28 & 0.47 \\
 & LS & 2.37 & 4.50 & 0.66 & 3.58 & 4.00 & 0.99 & 0.23 & 1.73 & 0.23 & 1.10 & 3.00 & 0.61 \\
 & EAGLE3 & 3.15 & 1.54 & 0.88 & 3.87 & 1.80 & 1.06 & 1.65 & 1.80 & 1.65 & 1.65 & 1.70 & 0.98 \\
 & Cascade & 2.25 & 2.00 & 0.62 & 2.59 & 1.70 & 0.71 & 0.84 & 2.20 & 0.84 & 0.79 & 1.55 & 0.44 \\
 & \method & 6.27 & 4.40 & 1.74 & 9.93 & 4.38 & 2.74 & 2.87 & 6.90 & 2.87 & 2.34 & 3.17 & 1.39 \\
\multirow{6}{*}{MA} & Auto & 3.32 & 1.00 & 1.00 & 3.24 & 1.00 & 1.00 & 1.08 & 1.00 & 1.00 & 1.78 & 1.00 & 1.00 \\
 & ES & 3.04 & 3.17 & 0.91 & 2.42 & 2.57 & 0.74 & 1.02 & 4.20 & 0.95 & 1.36 & 2.00 & 0.77 \\
 & LS & 3.43 & 4.50 & 1.03 & 2.48 & 4.00 & 0.77 & 0.44 & 2.83 & 0.41 & 1.32 & 3.17 & 0.74 \\
 & EAGLE3 & 3.34 & 1.92 & 1.01 & 3.14 & 1.22 & 0.97 & 0.91 & 1.42 & 0.84 & 1.87 & 1.64 & 1.05 \\
 & Cascade & 2.00 & 1.67 & 0.60 & 2.67 & 1.70 & 0.82 & 0.68 & 2.00 & 0.63 & 0.98 & 1.58 & 0.55 \\
 & \method & 6.52 & 4.25 & 1.96 & 8.60 & 4.40 & 2.65 & 2.38 & 6.45 & 2.20 & 3.06 & 3.25 & 1.72 \\
\multirow{6}{*}{MT} & Auto & 2.50 & 1.00 & 1.00 & 2.43 & 1.00 & 1.00 & 1.06 & 1.00 & 1.00 & 1.60 & 1.00 & 1.00 \\
 & ES & 3.62 & 2.43 & 1.45 & 2.15 & 2.57 & 0.88 & 0.67 & 3.00 & 0.63 & 1.68 & 2.34 & 1.05 \\
 & LS & 2.21 & 3.40 & 0.89 & 3.01 & 4.50 & 1.24 & 0.24 & 1.31 & 0.22 & 1.39 & 3.17 & 1.01 \\
 & EAGLE3 & 2.30 & 1.42 & 0.92 & 3.01 & 1.90 & 1.24 & 1.01 & 1.64 & 0.95 & 1.28 & 1.00 & 0.80 \\
 & Cascade & 1.31 & 1.54 & 0.52 & 1.52 & 1.75 & 0.63 & 0.84 & 1.91 & 0.80 & 0.84 & 1.50 & 0.61 \\
 & \method & 4.41 & 3.00 & 1.76 & 4.95 & 3.00 & 2.04 & 1.77 & 4.67 & 1.67 & 2.76 & 3.94 & 1.73 \\
\multirow{6}{*}{QA} & Auto & 3.84 & 1.00 & 1.00 & 2.57 & 1.00 & 1.00 & 0.96 & 1.00 & 1.00 & 1.71 & 1.00 & 1.00 \\
 & ES & 2.07 & 2.25 & 0.54 & 1.95 & 2.50 & 0.76 & 1.93 & 5.67 & 2.01 & 1.52 & 2.57 & 0.89 \\
 & LS & 2.72 & 4.50 & 0.71 & 2.15 & 3.60 & 0.83 & 0.50 & 3.00 & 0.52 & 1.30 & 3.17 & 0.76 \\
 & EAGLE3 & 2.12 & 1.54 & 0.55 & 2.27 & 1.31 & 0.88 & 1.29 & 2.00 & 1.34 & 1.67 & 1.70 & 0.98 \\
 & Cascade & 2.11 & 1.80 & 0.55 & 2.03 & 1.42 & 0.79 & 0.62 & 1.70 & 0.64 & 1.43 & 1.64 & 0.83 \\
 & \method & 5.74 & 3.60 & 1.49 & 5.72 & 3.78 & 2.22 & 2.34 & 5.67 & 2.44 & 2.78 & 3.00 & 1.63 \\
\multirow{6}{*}{RG} & Auto & 3.78 & 1.00 & 1.00 & 3.00 & 1.00 & 1.00 & 0.98 & 1.00 & 1.00 & 1.63 & 1.00 & 1.00 \\
 & ES & 2.47 & 2.55 & 0.65 & 3.23 & 4.75 & 1.08 & 0.98 & 4.50 & 1.00 & 1.39 & 2.00 & 0.85 \\
 & LS & 2.10 & 4.00 & 0.56 & 3.52 & 5.50 & 1.17 & 0.34 & 2.56 & 0.35 & 1.43 & 3.17 & 0.91 \\
 & EAGLE3 & 3.08 & 1.44 & 0.81 & 3.21 & 2.00 & 1.07 & 1.05 & 1.31 & 1.08 & 1.18 & 1.00 & 0.72 \\
 & Cascade & 1.96 & 1.73 & 0.52 & 1.36 & 1.82 & 0.45 & 0.70 & 1.90 & 0.71 & 1.02 & 1.58 & 0.65 \\
 & \method & 4.61 & 2.87 & 1.22 & 4.06 & 2.83 & 1.35 & 1.53 & 4.12 & 1.56 & 2.41 & 2.57 & 1.48 \\
\multirow{6}{*}{SU} & Auto & 2.50 & 1.00 & 1.00 & 2.39 & 1.00 & 1.00 & 1.01 & 1.00 & 1.00 & 1.77 & 1.00 & 1.00 \\
 & ES & 2.85 & 3.33 & 1.14 & 1.91 & 2.57 & 0.80 & 0.82 & 3.33 & 0.81 & 1.33 & 2.12 & 0.75 \\
 & LS & 2.69 & 4.50 & 1.08 & 2.09 & 4.00 & 0.87 & 0.31 & 1.80 & 0.31 & 1.37 & 3.17 & 0.78 \\
 & EAGLE3 & 1.43 & 1.21 & 0.57 & 1.85 & 1.31 & 0.78 & 1.04 & 1.20 & 1.03 & 2.16 & 1.50 & 1.22 \\
 & Cascade & 1.36 & 1.70 & 0.55 & 1.29 & 1.82 & 0.54 & 0.43 & 1.70 & 0.43 & 0.75 & 1.42 & 0.43 \\
 & \method & 4.50 & 3.53 & 1.80 & 3.47 & 2.25 & 1.45 & 1.21 & 3.33 & 1.20 & 2.67 & 3.00 & 1.51 \\
\multirow{6}{*}{TR} & Auto & 2.68 & 1.00 & 1.00 & 2.26 & 1.00 & 1.00 & 0.99 & 1.00 & 1.00 & 1.92 & 1.00 & 1.00 \\
 & ES & 1.90 & 1.80 & 0.71 & 1.95 & 1.89 & 0.86 & 0.98 & 3.50 & 0.99 & 1.66 & 2.71 & 0.87 \\
 & LS & 2.10 & 4.60 & 0.78 & 3.17 & 4.50 & 1.40 & 0.26 & 1.31 & 0.26 & 1.29 & 3.17 & 0.67 \\
 & EAGLE3 & 2.26 & 1.42 & 0.84 & 2.38 & 1.39 & 1.05 & 0.84 & 1.46 & 0.85 & 1.82 & 1.54 & 0.95 \\
 & Cascade & 1.43 & 1.42 & 0.54 & 2.83 & 1.80 & 1.25 & 0.73 & 2.10 & 0.74 & 1.22 & 1.80 & 0.64 \\
 & \method & 3.96 & 2.87 & 1.48 & 6.39 & 3.82 & 2.83 & 1.42 & 3.72 & 1.44 & 2.59 & 2.86 & 1.35 \\
\bottomrule
\end{tabular}%
}
\end{minipage}
\end{table*}

\clearpage

\newpage
\bibliographystyle{ACM-Reference-Format}
\bibliography{main}

\end{document}